\RequirePackage{fix-cm}
\documentclass[twocolumn,natbib,final]{svjour3}          
\smartqed  
\usepackage{graphicx}
\usepackage{amsmath}
\usepackage{amssymb}
\usepackage{overpic}
\usepackage[caption=false]{subfig}
\usepackage{hyperref}
\usepackage{xcolor}
\journalname{Autonomous Robots}
\begin{document}

\title{Learning rewards from exploratory demonstrations using probabilistic temporal ranking}


\titlerunning{Probabilistic temporal ranking}        

\author{Michael Burke, Katie Lu, Daniel Angelov,  Art\={u}ras Strai\v{z}ys, Craig Innes, Kartic Subr and Subramanian Ramamoorthy}

\authorrunning{Burke et. al} 

\institute{M. Burke \at
              Electrical and Computer Systems Engineering\\ Monash University, Australia \\
              \email{michael.g.burke@monash.edu}           %
           \and
           K. Lu, D. Angelov,  A. Strai\v{z}ys, C. Innes, K. Subr and S. Ramamoorthy \at
              School of Informatics\\ 
              University of Edinburgh, UK\\
              \email{s.ramamoorthy@ed.ac.uk} 
}

\date{Received: 22 Feb 2021 / Accepted: date}

\maketitle

\begin{abstract}
Informative path-planning is a well established approach to visual-servoing and active viewpoint selection in robotics, but typically assumes that a suitable cost function or goal state is known. This work considers the inverse problem, where the goal of the task is unknown, and a reward function needs to be inferred from exploratory example demonstrations provided by a demonstrator, for use in a downstream informative path-planning policy. Unfortunately, many existing reward inference strategies are unsuited to this class of problems, due to the exploratory nature of the demonstrations. In this paper, we propose an alternative approach to cope with the class of problems where these sub-optimal, exploratory demonstrations occur. We hypothesise that, in tasks which require discovery, successive states of any demonstration are progressively more likely to be associated with a higher reward, and use this hypothesis to generate time-based binary comparison outcomes and infer reward functions that support these ranks, under a probabilistic generative  model. We formalise this \emph{probabilistic temporal ranking} approach and show that it improves upon existing approaches to perform reward inference for autonomous ultrasound scanning, a novel application of learning from demonstration in medical imaging while also being of value across a broad range of goal-oriented learning from demonstration tasks.
\end{abstract}

\section{Introduction}

Informative path-planning for visual servo control and active viewpoint selection is a key ability for modern day autonomous robotics. However, these approaches typically assume that some notion of a goal or desired viewpoint is available, which may not always be the case. This work considers the case where the goal or cost function of an informative path-planning task is unknown, and needs to be inferred from expert demonstrations. The ability to teach robotic agents using expert demonstration of tasks promises exciting developments across several sectors of industry. This is particularly true of medical imaging, where task and anatomy variability can make it challenging to provide specifications and describe tasks directly, and it may be more natural to consider an apprenticeship learning \citep{abbeel2004apprenticeship} approach.

Indirect imitation learning \citep{Bagnell-2015-5921} approaches formulate apprenticeship learning as a search problem within a solution space of plans, where some notional (unknown) \emph{reward function} induces the demonstrated behaviour. A key learning problem is then to estimate this reward function. This \emph{reward inference} approach is commonly known as inverse reinforcement learning (IRL) \citep{ng2000algorithms}. 

\begin{figure}
    \centering
    \begin{minipage}{0.35\textwidth}
    \centering
     \subfloat[Demonstration traces]{\hspace{-4mm}\begin{tabular}{l}
     \includegraphics[width=0.98\textwidth]{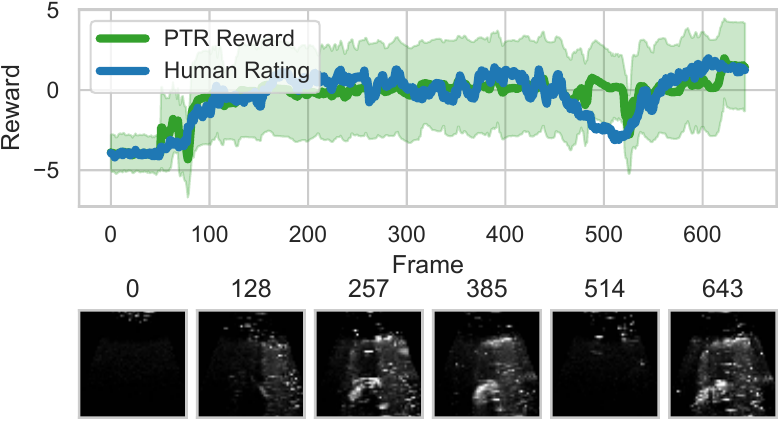}\\
     \includegraphics[width=0.98\textwidth]{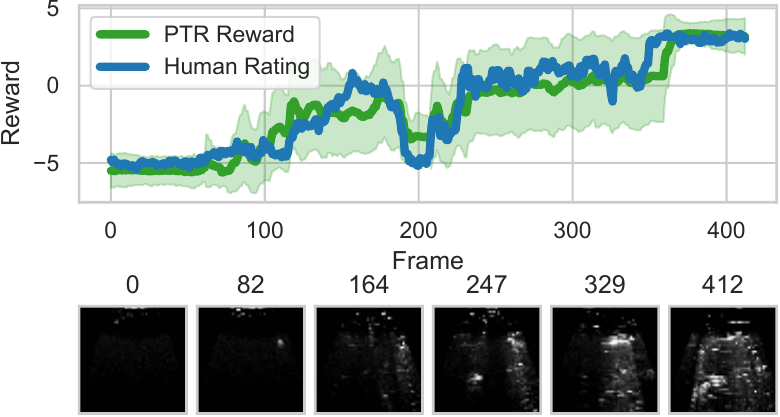}
     \end{tabular}}
     \end{minipage} \begin{minipage}{0.1\textwidth}
     \subfloat[Demo\label{fig:demo}]{\includegraphics[width=0.8\textwidth]{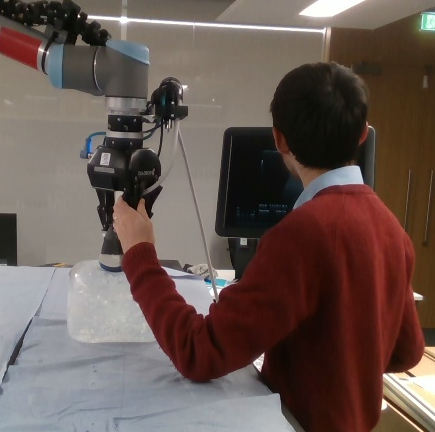}}\\
     \subfloat[Phantom]{\includegraphics[width=0.8\textwidth]{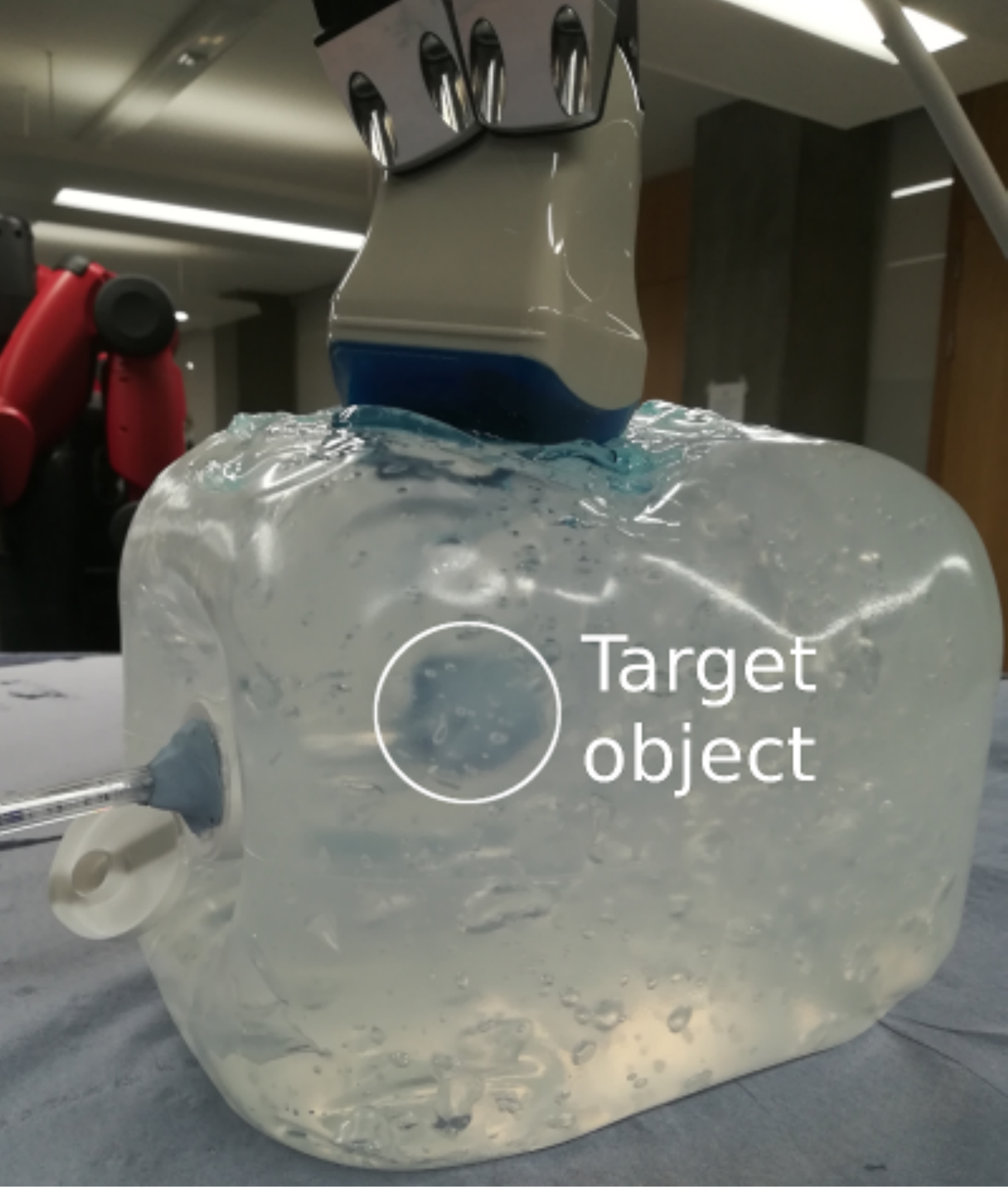}}\\
     \subfloat[Saliency\label{fig:saliency}]{\includegraphics[width=0.9\textwidth]{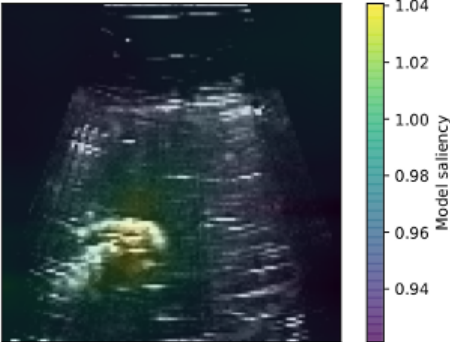}}
     \end{minipage}
    \caption{This work introduces a temporal ranking strategy to learn reward functions (a) from human demonstrations (b) for autonomous ultrasound scanning. Probabilistic temporal ranking can learn to identify non-monotonically increasing rewards from demonstration image sequences containing exploratory actions, and successfully associates ultrasound features corresponding to a target object (c) with rewards (d). \label{fig:unseen_reward}}
\end{figure}

However, as illustrated in the ultrasound scanning task of Figure \ref{fig:unseen_reward}, the demonstration process often followed by medical practitioners is a naturally exploratory one, involving an information gathering phase in addition to an optimisation phase, or a generally greedy motion towards a desired viewpoint. Human demonstrators regularly act so as to improve their states in complex tasks. For example, in a study on human approaches to combinatorial optimisation, \citet{murawski2016humans} found that humans behave relatively greedily, following something akin to a branch-and-bound algorithm for search. Similarly, in palpation experiments, \citet{Konstantinova13} showed that humans searching for hard excursions in soft tissue searched extensively for nodules, but acted greedily when these had been found, refining the search using small focused, circular movements. Unfortunately, the exploratory nature of demonstrations such as these poses a challenge for many existing approaches to reward inference. 

For example, the IRL approach to apprenticeship learning~\citep{abbeel2004apprenticeship} aims to match the frequency (counts) of features encountered in the learner’s behaviour with those observed in demonstrations. This technique provides necessary and sufficient conditions when the reward function is linear in the features encoding execution states, but results in ambiguities in associating optimal policies with reward functions or feature counts. An elegant reformulation of this using the principle of maximum entropy resolves ambiguities and results in a single optimal stochastic policy. Methods for maximum-entropy IRL \citep{ziebart2008maximum,wulfmeier2015maximum,levine2011nonlinear} identify reward functions using maximum likelihood estimation, typically under the assumption that the probability of seeing a given trajectory is proportional to the exponential of the total reward along a given path. Unfortunately, these methods are fundamentally frequentist and thus struggle to cope with \emph{repetitive sub-optimal demonstrations}, as they assume that frequent appearance implies relevance. i.e. If a feature is seen repeatedly across demonstration trajectories, it is deemed valuable, as are policies that result in observations of these features.

\begin{figure}
    \centering
    \includegraphics[height=0.25\textwidth]{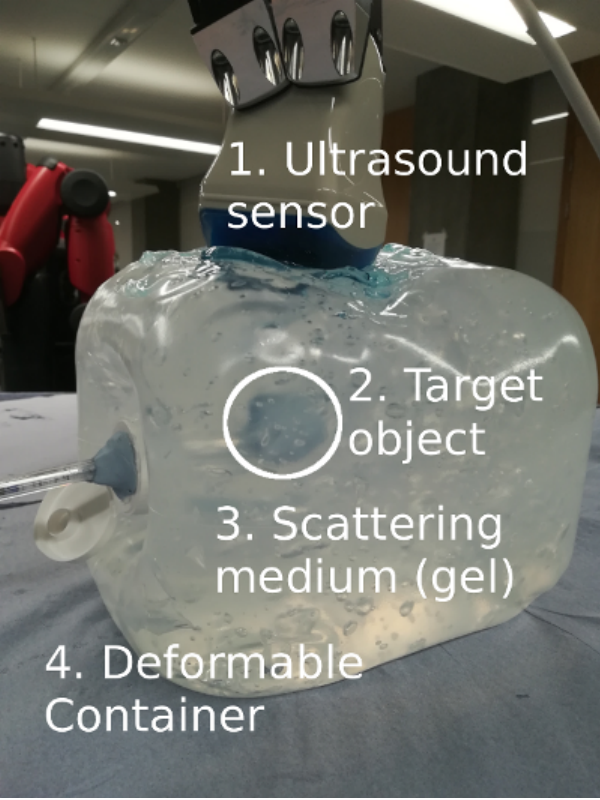}
    \includegraphics[height=0.25\textwidth]{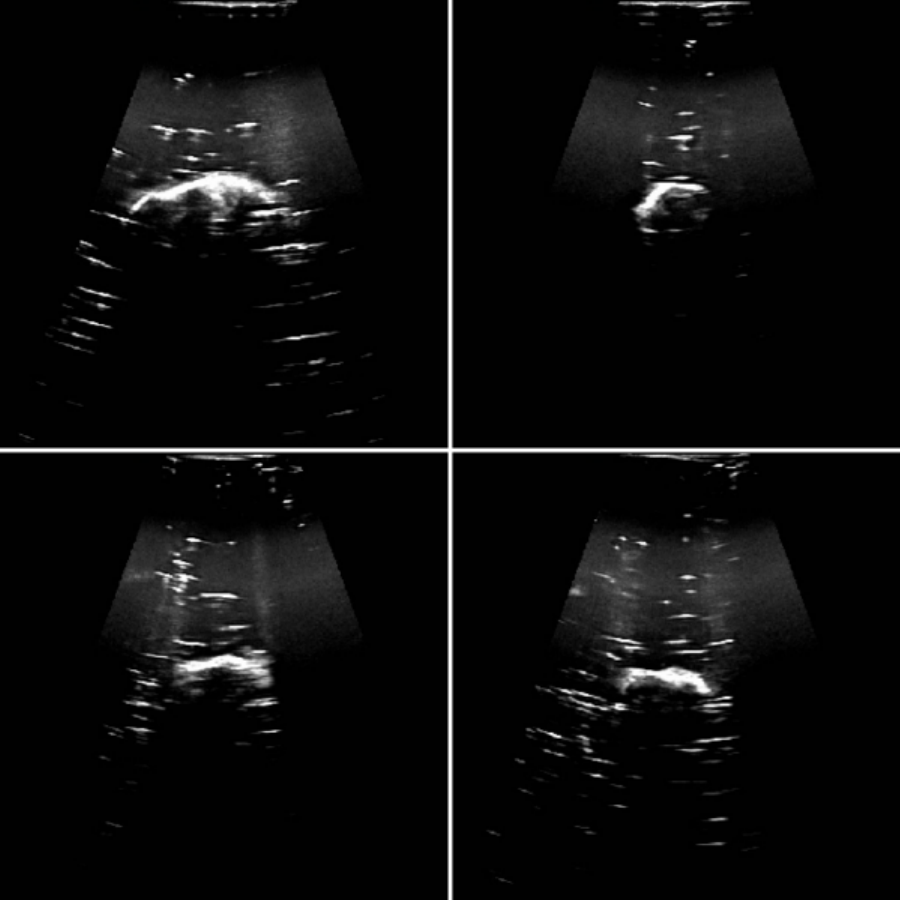}
    \caption{This work considers the task of learning to search for and capture an image of a target object (2.) suspended in a scattering material (3.) housed within a deformable container (4.). Our goal is to learn a reward signal from demonstrations that allows us to move an ultrasound sensor (1.) to positions that produce clear images of the target object. High quality ultrasound images (right) captured by a human demonstrator show high intensity contour outlines, centre the target object of interest, and generally provide some indication of target object size.\label{fig:expert}}
\end{figure}

This makes these approaches unsuitable for a broad class of tasks that require \emph{exploratory actions} or environment identification during demonstration.~e.g.~an expert using an ultrasound scan to locate a tumour (Figure \ref{fig:unseen_reward}). Obtaining useful ultrasound images requires contact with a deformable body (see Figure \ref{fig:expert}) at an appropriate position and contact force, with image quality affected by the amount of ultrasound gel between the body and the probe, and air pockets that obscure object detection. This means that human demonstrations are frequently and inherently sub-optimal, requiring that a demonstrator actively search for target objects, while attempting to locate a good viewpoint position and appropriate contact force. This class of demonstration violates many of the assumptions behind existing reward inference schemes.

In order to address this, this paper introduces probabilistic temporal ranking (PTR), a temporal ranking model of reward that addresses these limitations. PTR is a self-supervised approach and thus does not rely on the true reward or value function, using only a sequence of images or states to infer underlying reward functions.

Instead of assigning reward based on the maximum entropy model, PTR attributes reward using a ranking model. Here, we assume that, in general, an expert acts to improve their current state. This means that it is likely that observations at a later stage in a demonstrated trajectory are more important than those seen at an earlier stage. PTR uses this fact to generate time-based binary comparison outcomes, and then uses these to infer reward functions that support these ranks, under a probabilistic generative model that combines  information about image or observation similarity (a Gaussian process reward model over a learned latent space) with a noisy pairwise  generative  outcome model.

Importantly, PTR is able to handle cases where this temporal improvement is unsteady and non-monotonic, with intermediate performance dips. Experimental results show that probabilistic temporal ranking successfully recovers reward maps from demonstrations in tasks requiring significant levels of exploration alongside exploitation (where maximum entropy IRL fails), and obtains similar performance to maximum entropy inverse IRL when optimal demonstrations are available.

PTR is not only useful for active viewpoint selection problems, but the assumptions governing generally increasing rewards hold for a broad class of goal-oriented control problems, for example, swinging up a pendulum, or navigating to a specific location.

We highlight this through a number of simulated experiments, and illustrate the value of our approach in a challenging ultrasound scanning application, where demonstrations inherently contain a searching process, and show that we can train a model to find a tumour-like mass in an imaging phantom\footnote{An imaging phantom is an object that mimics the physical responses of biological tissue, and is commonly used in medical imaging to evaluate and analyse imaging devices.}. Ultrasound imaging is a safe and low cost sensing modality of significant promise for surgical robotics, and is already frequently used for autonomous needle steering and tracking \citep{LIANG2010173,6630795}. \citet{chatelain2015optimization} propose the use of ultrasound quality maps to improve image quality in robotic ultrasound scanning applications. Autonomous visual servoing systems have also been proposed in support of teleoperated ultrasound diagnosis \citep{988970,6224974}, but these techniques tend to rely on hand designed anatomical target detectors or confidence maps.  The scanner introduced in this work is fully autonomous, and relies entirely on a reward signal learned from demonstration, in what we believe is a first for medical imaging. Importantly, the probabilistic temporal ranking formulation provides more signal for learning, as a greater number of comparisons can be generated from each demonstration trajectory. This means that we can train a more effective prediction model from pixels than with maximum entropy IRL, which in turn opens up a number of avenues towards self-supervised learning for medical imaging and diagnosis.

In summary, the primary contributions of this paper are
\begin{itemize}
    \item a temporal ranking reward model that allows for reward inference from sub-optimal, high dimensional exploratory demonstrations, and
    \item a method for autonomous ultrasound scanning using image sequence demonstrations. 
\end{itemize}

\section{Related work}

\subsection{Reward or cost function inference}

As mentioned previously, apprenticeship learning \citep{abbeel2004apprenticeship} is an alternative to direct methods of imitation learning \citep{Bagnell-2015-5921} or behaviour cloning, and is currently dominated by indirect approaches making use of maximum entropy assumptions.

Maximum entropy or maximum likelihood inverse reinforcement learning models the probability of a user preference for a given trajectory $\zeta$ as proportional to the exponential of the total reward along the path \citep{ziebart2008maximum},
\begin{equation}
    p(\zeta|r) \propto \exp(\sum_{s,a\in \zeta}r_{s,a}).
\end{equation}
Here, $s$ denotes a state, $a$ an action, and $r_{s,a}$ the reward obtained for taking an action in a given state. It is clear that this reward model can be maximised by any number of reward functions. \citet{levine2011nonlinear} use a Gaussian process prior to constrain the reward, while \citet{wulfmeier2015maximum} backpropagate directly through the reward function using a deep neural network prior. Maximum entropy inverse reinforcement learning approaches are typically framed as iterative policy search, where policies are identified to maximise the reward model. This allows for the incorporation of additional policy constraints and inductive biases towards desirable behaviours, as in relative entropy search \citep{boularias2011relative}, which uses a relative entropy term to keep policies near a baseline, while maximising reward feature counts. Maximum entropy policies can also be obtained directly, bypassing reward inference stages, using adversarial imitation learning \citep{ho2016generative,finn2016connection, fu2017learning,ghasemipour2019divergence}, although reward prediction is itself useful for medical imaging applications.

Although maximum entropy IRL is ubiquitous, alternative reward models have been proposed. For example, \citet{angelov2019composing} train a neural reward model using demonstration sequences, to schedule high level policies in long horizon tasks. Here, they capture overhead scene images, and train a network to predict a number between $0$ and $1$, assigned in increasing order to each image in a demonstration sequence. This ranking approach is similar to the pairwise ranking method we propose, but, as will be shown in later results, is limited by its rigid assumption of linearly increasing reward. \citet{Majumdar-RSS-17} propose flexible reward models that explicitly account for human risk sensitivity. Time contrastive networks \citep{sermanet2018time} learn disentangled latent representations of video using time as a supervisory signal. Here, time synchronised images taken from multiple viewpoints are used to learn a latent embedding space where similar images (captured from different viewpoints) are close to one another. This embedding space can then be used to find policies from demonstrations. Time contrastive networks use a triplet ranking loss, and are trained using positive and negative samples (based on frame timing margins).

Preference-based ranking of this form is widely used in inverse reinforcement learning to rate demonstrations \citep{wirth2017survey}, and preference elicitation \citep{braziunas2006preference} is a well established area of research. For example, \citet{brochu2010bayesian} use Bayesian optimisation with a pairwise ranking model to allow users to procedurally generate realistic animations. \citet{active_preference_learning,Biyik-RSS-20} and \cite{tucker2020preference} actively query demonstrators to learn reward functions. The latter make use of a Thurstonian model \cite{thurstone2017law}, computing the cumulative distribution function over the difference between rewards. This is intractable, and requires a Laplace approximation for inference.   \citet{sugiyama2012preference} use preference-based inverse reinforcement learning for dialog control. Here, dialog samples are annotated with ratings, which are used to train a preference-based reward model. These preference elicitation approaches are effective, but place a substantial labelling burden on users. In this work, we consider the non-interactive learning case where we are required to learn directly from unlabelled observation traces.


A number of extensions to maximum entropy IRL have been proposed to cope with sub-optimal demonstration sequences, typically through the inclusion of additional supervisory information about the quality of a demonstration sequence \citep{wu2019imitation,brown2019extrapolating}. In large part, these works define demonstration quality in terms of how noisy they are \citep{brown2019extrapolating} or how far they deviate from some perfect demonstration. However, for discovery tasks such as ultrasound scanning, it is much harder to determine what constitutes an optimal or perfect demonstration, as all demonstrations require a degree of exploration before finding a good viewpoint. 

\citet{lee2016inverse} use leveraged Gaussian processes to learn from both positive and negative demonstration examples. Similarly, \citet{shiarlis:aamas16} and \citet{valko2013semi} consider inverse reinforcement learning in the looser case where only a subset of demonstrations are considered expert or successful, and the remaining `failures' may contain elements key to success, or even be unlabelled successful demonstrations. These are semi-supervised learning approaches, as they rely on additional labelling information about the quality of demonstrations. In contrast, the PTR approach proposed in this paper is self-supervising, as it relies on time as a supervisory signal.   

\citet{brown2020better} make use of a preference ranking approach to improve robot policies through artificial trajectory ranking using increasing levels of injected noise. Unlike \citet{brown2020better}, which uses preference ranking over trajectories, our work uses preference ranking within trajectories, under the assumption that a demonstrator generally acts to improve or maintain their current state. We modify a Bayesian image ranking model \citep{burke2017rapid} that accounts for potential uncertainty in this assumption, and is less restrictive than the linearly increasing model of \citet{angelov2019composing}. Bayesian ranking models \citep{chu2005preference} are common in other fields -- for example, TrueSkill\textsuperscript{TM} \citep{herbrich2007trueskill} is widely used for player performance modelling in online gaming settings, but has also been applied to to train image-based style classifiers in fashion applications \citep{kiapour2014hipster} and to predict the perceived safety of street scenes using binary answers to the question ``Which place looks safer?'' \citep{naik2014streetscore}. 

\subsection{Active viewpoint selection}

Given an appropriate reward model, autonomous ultrasound scanning requires a policy that balances both exploration and exploitation for active viewpoint selection or informative path planning. Research on active viewpoint selection \citep{SRIDHARAN2010704} is concerned with agents that choose viewpoints which optimise the quality of the visual information they sense. Similarly, informative path planning involves an agent choosing actions that lead to observations which most decrease uncertainty in a model. Gaussian processes (GP) are frequently used for informative path planning because of their inclusion of uncertainty, data-efficiency, and flexibility as non-parametric models. 

\citet{binney2012branch} use GPs with a branch and bound algorithm, while \citet{cho2018informative} perform informative path planning using GP regression and a mutual information action selection criterion. More general applications of GPs to control include PILCO \citep{deisenroth2011pilco}, where models are optimised to learn policies for reinforcement learning control tasks, and the work of \citet{ling2016gaussian}, which introduces a GP planning framework that uses GP predictions in H-stage Bellman equations.

These Bayesian optimisation schemes are well established methods for optimisation of an unknown function, and have been applied to many problems in robotics including policy search \citep{7989380}, object grasping \citep{yi2016active}, and bipedal locomotion \citep{calandra2014}. 

By generating policies dependent on predictions for both reward value and model uncertainty, Bayesian optimisation provides a mechanism for making control decisions that can both progress towards some task objective and acquire information to reduce uncertainty. GP’s and Bayesian optimisation are often used together, with a GP acting as the surrogate model for a Bayesian optimisation planner, as in the mobile robot path planning approaches of \citet{martinez2009bayesian} and \citet{6907763}. Our work takes a similar approach, using GP-based Bayesian optimisation for path planning in conjunction with the proposed observation ranking reward model.

The combination of preference-based learning with a policy trading-off exploration-exploitation is commonly studied within duelling bandit frameworks \citep{sui2017multi}. Here, instead of learning from a reward signal, a policy is required to learn directly from preferential feedback. This differs from the the reward inference setting studied in this paper, where preference signals are generated a-priori by generating temporal comparisons from human demonstrations, and not online when the policy is deployed.

\section{Probabilistic temporal ranking}

This paper introduces probabilistic temporal ranking (PTR), a reward inference strategy for high dimensional exploratory image demonstrations. Below, we first describe a fully probabilistic temporal ranking model, which we then examine using a series of simulated experiments. We then introduce a deterministic neural approximation that can be efficiently trained in a fully end-to-end fashion, and is more suited to larger training sets, before moving on to our primary autonomous ultrasound scanning experiments, and the description of a Bayesian optimisation strategy for informative path-planning that facilitates this. 

\begin{figure*}
    \centering
    \includegraphics[width=\textwidth]{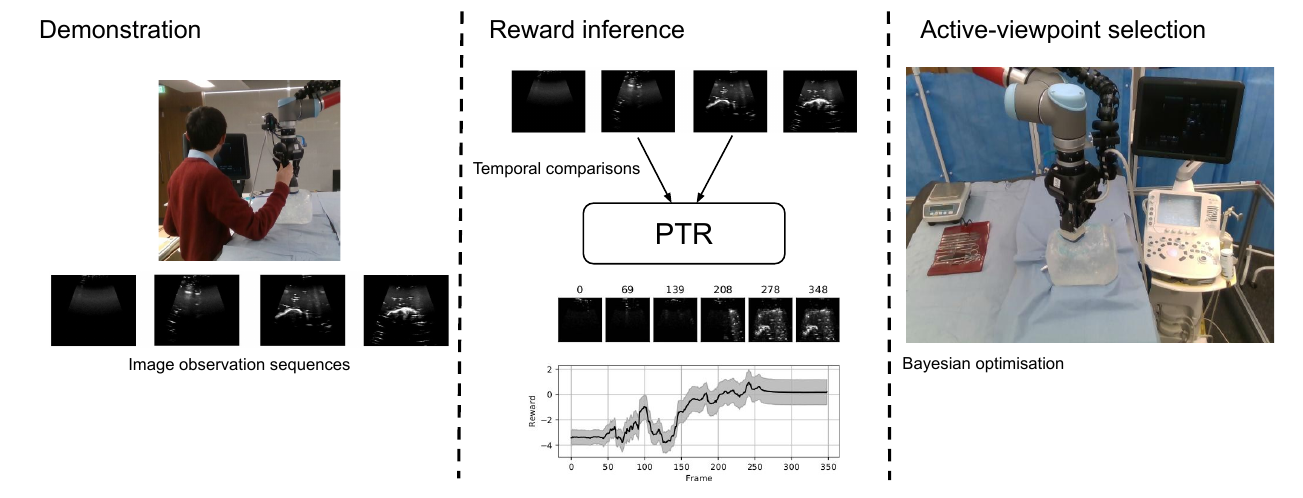}
    \caption{Pipeline for autonomous ultrasound scanning. User demonstrations are used to collect image sequences, temporal comparisons are then sampled from these sequences and used by PTR to train a reward inference model, which is then used by a Bayesian optimisation policy for active-viewpoint selection in new environments.}
    \label{fig:pipeline}
\end{figure*}

In general, we envisage PTR being used as in Figure \ref{fig:pipeline}. First, a series of observations are collected while a human demonstrates an exploratory visual scanning task. PTR is then used to train a reward model by sampling pairwise temporal comparison outcomes from the demonstration sequences and performing model fitting. This reward model is then used by a suitable informative path-planning or active viewpoint selection policy (in this case Bayesian optimisation) to replicate the demonstration in new environments.

\subsection{Fully probabilistic model}

This paper incorporates additional assumptions around the structure of demonstration sequences, to allow for improved reward inference. We introduce a reward model that learns from pairwise comparisons sampled from demonstration trajectories. Here we assume that an observation or state seen later in a demonstration trajectory should typically generate greater reward than one seen at an earlier stage.

We build on the pairwise image ranking model of \citet{burke2017rapid}, replacing pre-trained object recognition image features with a latent state, $\mathbf{x}_t \in \mathbf{R}^d$, learned using a convolutional variational autoencoder (CVAE),
\begin{equation}
\mathbf{x}_t \sim \mathcal{N}(\mu(\mathbf{Z}_t),\sigma(\mathbf{Z}_t)),
\end{equation}
that predicts mean, $\mu(\mathbf{Z}_t) \in \mathbb{R}^d$, and diagonal covariance, $\sigma(\mathbf{Z}_t)\in \mathbb{R}^{d \times d}$, for input observation $\mathbf{Z}_t\in \mathbb{R}^{w \times h}$ captured at time $t$ (assuming image inputs of dimension $ w \times h$).

Rewards $r_t \in \mathbb{R}^1$ are modelled using a Gaussian process prior,
\begin{equation}
\begin{bmatrix} r' \\ r_t\end{bmatrix} \sim \mathcal{N}\left(\mathbf{0},\begin{bmatrix}K(\mathbf{X}',\mathbf{X}')+\Sigma_n & K(\mathbf{X}',\mathbf{x}_t)\\K(\mathbf{x}_t,\mathbf{X}') & K(\mathbf{x}_t,\mathbf{x}_t)\end{bmatrix}\right). \label{eq:Gauss}
\end{equation}
Here, we use $\mathbf{X}'$ and $r'$ to denote states and reward pairs corresponding to training observations. $\mathbf{X}' \in \mathbb{R}^{N\times d}$ is a matrix formed by vertically stacking $N$ latent training states, and $K(\mathbf{X}',\mathbf{X}')$ a covariance matrix formed by evaluating a Matern32 kernel function
\begin{equation}
k(\mathbf{x}_t,\mathbf{y}_t) = \mathrm{Matern32}(\mathbf{x}_t,\mathbf{y}_t,l), \;\; \mathbf{x}_t, \mathbf{y}_t \in \mathbb{R}^d\\
\end{equation}
for all possible combinations of latent state pairs $\mathbf{x}_t ,\mathbf{y}_t$, sampled from the rows of $\mathbf{X}'$. $l \in \mathbb{R}^1$ is a length scale parameter with a Gamma distributed prior, $l \sim \Gamma(\alpha=2.0,\beta=0.5)$, and $\Sigma_n \in \mathbb{R}^{N \times N}$ is a diagonal heteroscedastic noise covariance matrix, with diagonal elements drawn from a Half Cauchy prior, $\Sigma_n \sim \text{HalfCauchy}(\beta=1.0)$. 

These priors are well calibrated to the inference task here, and should not need to be adjusted in other applications. A half Cauchy prior $(\beta=1)$ is a heavy tailed distribution that allows for diagonal covariance parameters (see Appendix, Figure \ref{fig:Hypers}), favouring low noise rewards, but also allowing for higher noise if needed. Inference with this prior is thus capable of handling both noisy and more repeatable rewards. Decreasing beta would increase the prior probability of little variability in rewards for a given state.

Similarly, the Gamma distributed length scale prior places most probability mass over a length scale of about 1, but allows a range of values (see Appendix, Figure \ref{fig:Hypers}). Given the standard normal prior used by the variational autoencoder, which compresses observations into a state space roughly constrained within the range (-3, 3), this prior allows for both small local influence between latent states and rewards, or wider correspondences across the latent space if needed.

At prediction time, reward predictions $r_t$ for image observations $\mathbf{Z}_t$ can be made by encoding the image to produce latent state $\mathbf{x}_t$, and conditioning the Gaussian in Equation (\ref{eq:Gauss}) \citep{williams2006gaussian}.

Using this model, the generative process for a pairwise comparison outcome, $g \in \{0,1\}$, between two input observation rewards $r_{t_1}$ and $r_{t_2}$ at time steps $t_1$ and $t_2$, is modelled using a Bernoulli trial over the sigmoid of the difference between the rewards,
\begin{equation}
g \sim \text{Ber}\left(\text{Sig}(r_{t_2}-r_{t_1})\right).
\end{equation}
This Bernoulli trial introduces slack in the model, allowing for tied or even decreasing rewards to be present in the demonstration sequence. 

The Sigmoid used here produces a logit and allows for simple differentiation, which is helpful for approximate inference schemes and the neural approximation introduced below, avoiding the need for the Laplace approximation to the posterior used in \cite{tucker2020preference,Biyik-RSS-20}.

\subsection{Reward inference using temporal observation ranking}

The generative model above is fit to demonstration sequences using automatic differentiation variational inference (ADVI) \citep{kucukelbir2017automatic} by sampling N observation pairs $\mathbf{Z}_{t_1},\mathbf{Z}_{t_2}$ from each demonstration sequence, which produce a comparison outcome
\begin{equation}
g = \begin{cases} 1 &  \text{if}\,\,\, t_2 \geq t_1\\ $0$ & \text{if}\,\,\, t_1 < t_2\end{cases}.
\end{equation}
\begin{figure}
    \centering
    \includegraphics[width=0.48\textwidth]{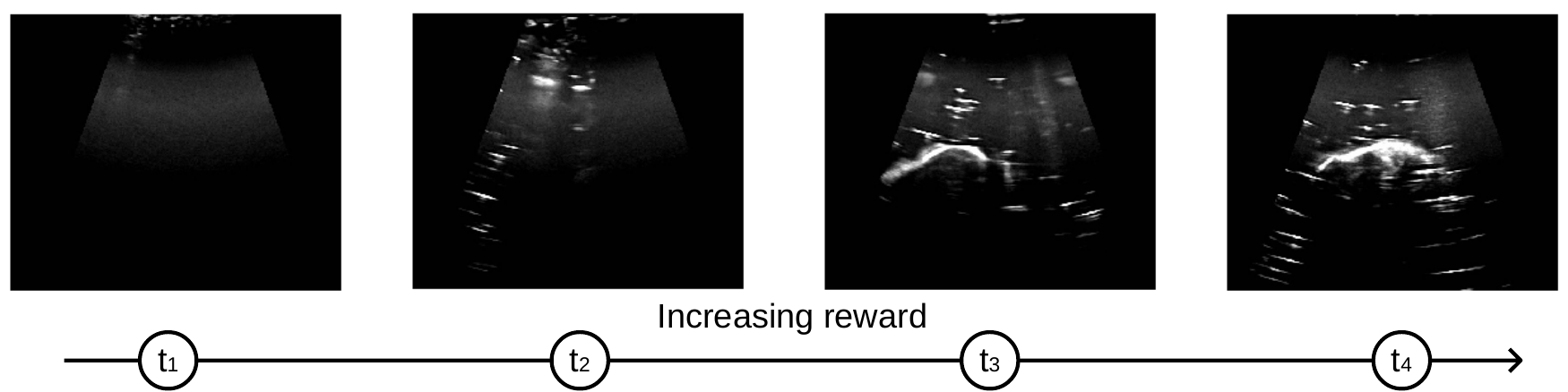}
    \caption{Time is used as a supervisory signal, by sampling image pairs at times $t_i, t_j$, and setting $g=1 \; \mathrm{if} \; t_i > t_j, g=0$ otherwise.\label{fig:timeline}}
\end{figure}

Intuitively, this temporal comparison test, which uses time as a supervisory signal (Figure ~\ref{fig:timeline}), operates as follows. Assume that an image captured at time step $t_2$ has greater reward than an image captured at $t_1$. This means that the sigmoid of the difference between the rewards is likely to be greater than 0.5, which leads to a higher probability of returning a comparison outcome $g=1$.  Importantly, this Bernoulli trial allows some slack in the model -- when the difference between the rewards is closer to 0.5, there is a greater chance that a comparison outcome of $g=1$ is generated by accident. This means that the proposed ranking model can deal with demonstration trajectories where the reward is non-monotonic. Additional slack in the model is obtained through the heteroscedastic noise model, $\Sigma_n$, which also allows for uncertainty in inferred rewards to be modelled.

Inference under this model amounts to using the sampled comparison outcomes from a demonstration trajectory to find rewards that generate similar comparison outcomes, subject to the Gaussian process constraint that images with similar appearance should exhibit similar rewards. After inference, we make reward predictions by encoding an input image, and evaluating the conditional Gaussian process at this latent state.

We briefly illustrate the value of this probabilistic temporal ranking approach in exploratory tasks using two simple grid world experiments.

\subsection{Grid world -- optimal demonstrations}

The first experiment considers a simple grid world, where a Gaussian point attractor is positioned at some unknown location. Our goal is to learn a reward model that allows an agent (capable of moving up, down, left and right) to move towards the target location. For these experiments, our state is the agent's 2D grid location.

\begin{figure}
    \centering
    \subfloat[Ground truth]{
    \includegraphics[width=0.23\textwidth]{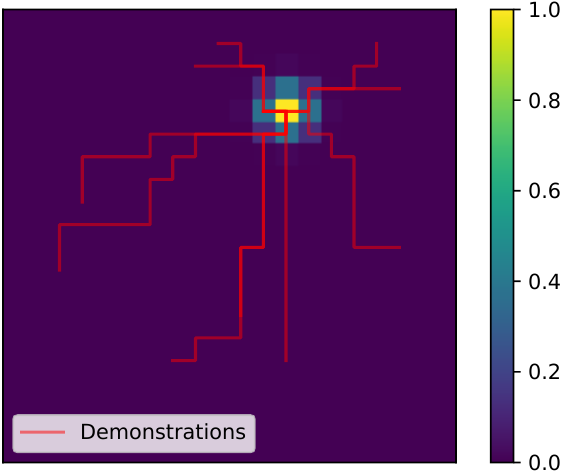}
    }
    \subfloat[GP-ME-IRL]{
    \includegraphics[width=0.23\textwidth]{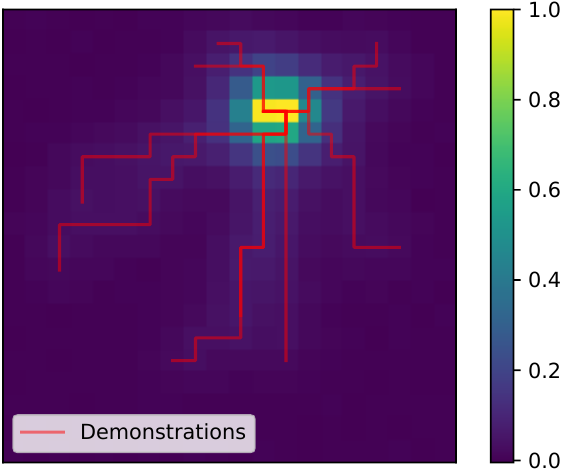}
    }\\
    \subfloat[GP-PTR]{
    \includegraphics[width=0.23\textwidth]{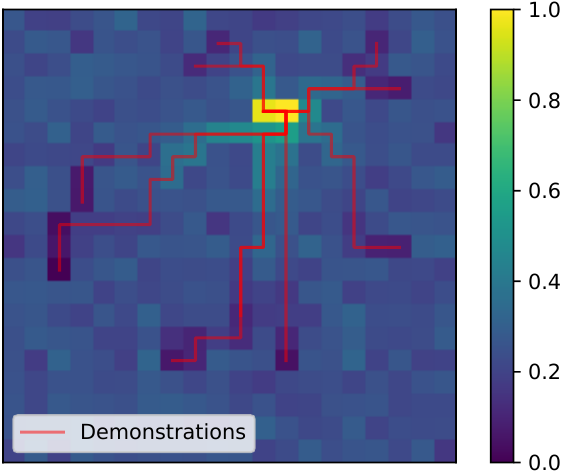}}
    \subfloat[GP-LTR]{
    \includegraphics[width=0.23\textwidth]{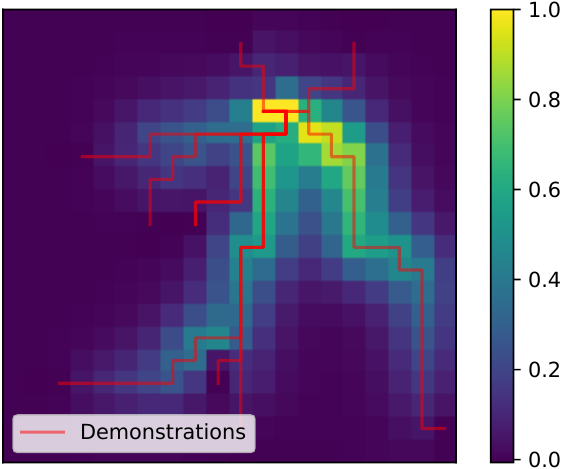}
    }
    \caption{Reward inference from \textbf{optimal} demonstrations. Demonstration trajectories are marked in red, and the colour map indicates the reward for each grid position. PTR and ME models have similar relative reward values, and policies trained using these rewards perform near identically. A linearly increasing reward model (LTR) attributes reward more evenly across a demonstration, resulting in sub-optimal policy performance here.\label{fig:optimal}}
\end{figure}

We generate 5 demonstrations (grid positions) from random starting points, across 100 randomised environment configurations with different goal points. We then evaluate performance over 100 trials in each configuration, using a policy obtained through tabular value iteration (VI) using the reward model inferred from the 5 demonstrations. This policy is optimal, as the target location is known, so for all demonstrations the agent moves directly towards the goal, as illustrated for the sample environment configuration depicted in Figure \ref{fig:optimal}.

Table \ref{tab:optimal_grid} shows the averaged total returns obtained for trials in environments when rewards are inferred from optimal demonstrations using the probabilistic temporal ranking\footnote{We use PyMC3 \citep{10.7717/peerj-cs.55} (GP-PTR) to build probabilistic reward models and ADVI \citep{kucukelbir2017automatic} for model fitting.} (GP-PTR), a Gaussian process maximum entropy approach \citep{levine2011nonlinear} (GP-ME-IRL)  and an increasing linear model assumption \citep{angelov2019composing} (GP-LTR). Value iteration is used to find a policy using the mean inferred rewards.
\begin{table}
    \centering
    \caption{Averaged total returns using VI policy trained using inferred reward from optimal demonstrations.}
    \label{tab:optimal_grid}
    \begin{tabular}{c|c}
    \hline
        & Reward \\
        \hline
         GP-PTR & 9.51 $\pm$ 4.92\\
        \hline
         GP-ME-IRL & 9.58 $\pm$ 4.90\\
         \hline
         GP-LTR & 7.39 $\pm$ 5.72\\
         \hline
    \end{tabular}
\end{table}

In the optimal demonstration case, policies obtained using both the maximum entropy and probabilistic temporal ranking approach perform equally well, although PTR assigns more neutral rewards to unseen states (Figure \ref{fig:optimal}). Importantly, as the proposed model is probabilistic, the uncertainty in predicted reward can be used to restrict a policy to regions of greater certainty by performing value iteration using an appropriate acquisition function instead of the mean reward. This implicitly allows for risk-based policies -- by weighting uncertainty higher, we could negate the neutrality of the ranking model (risk-averse). Alternatively, we could tune the weighting to actively seek out uncertain regions with perceived high reward (risk-seeking). 

\begin{figure}
    \centering
    \subfloat[Ground truth]{
        \includegraphics[width=0.23\textwidth]{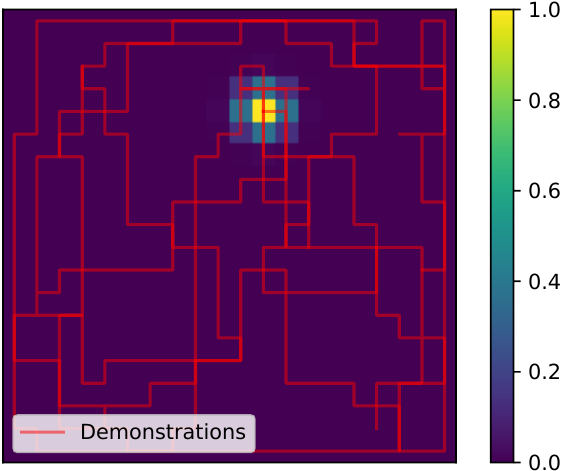}
        }
   \subfloat[GP-ME-IRL]{
    \includegraphics[width=0.23\textwidth]{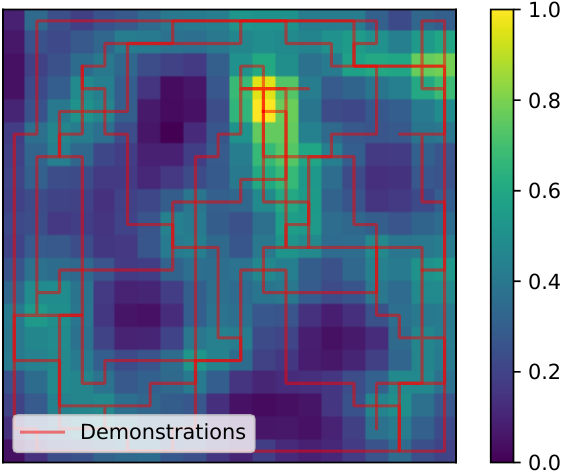}
    }\\
    \subfloat[GP-PTR]{  \includegraphics[width=0.23\textwidth]{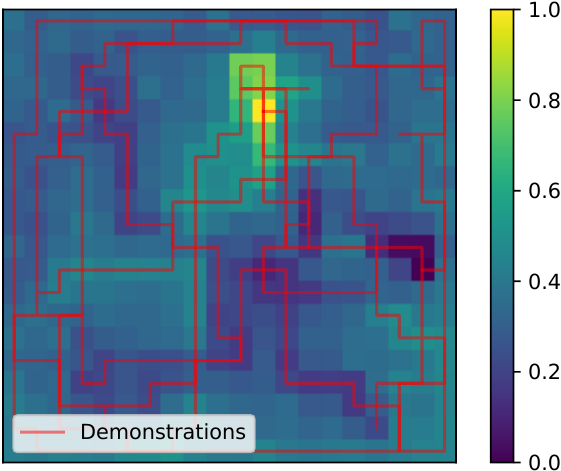}
    }
    \subfloat[GP-LTR]{
    \includegraphics[width=0.23\textwidth]{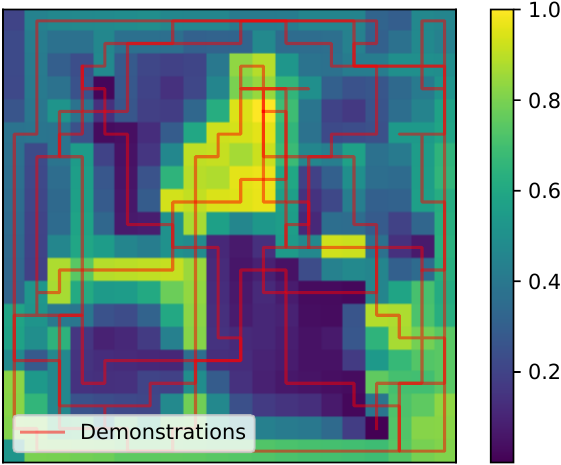}
    }
    \caption{Reward inference from \textbf{exploratory} demonstrations. Demonstration trajectories are marked in red, and the colour map indicates the reward for each grid position.\label{fig:suboptimal} Both the linearly increasing and maximum entropy reward models induce local maxima that result in sub-optimal policies.}
\end{figure}

\subsection{Grid world -- exploratory demonstrations}

Our second experiment uses demonstrations that are provided by an agent that first needs to explore the environment, before exploiting it. Here, we use a Gaussian process model predictive control policy (see below) to generate demonstrations, and repeat the experiments above. As shown in Figure~\ref{fig:suboptimal}, this policy may need to cover a substantial portion of the environment before locating the target.
 
Table \ref{tab:suboptimal_grid} shows the averaged total returns obtained for trials in environments when rewards are inferred from exploratory demonstrations using probabilistic temporal ranking, the Gaussian process maximum entropy approach and the linearly increasing reward assumption. Here, value iteration (VI) is used to find the optimal policy using the inferred rewards. A comparison with T-REX \citep{brown2019extrapolating} and D-REX \citep{brown2020better}, trajectory ranking methods designed to learn reward functions from sub-optimal demonstrations are also included. It should be noted that T-REX is a supervised learning method, relying on additional labelling information about the quality of a demonstration, while PTR requires no information beyond the demonstration sequences. For these experiments, we generate labelling information for T-REX by using trajectory length as a rough heuristic for the quality of a demonstration. D-REX generates ranked trajectories by artificially injecting noise to demonstrations.
\begin{table}
    \centering
    \caption{Averaged total returns using VI policy trained using inferred reward from exploratory demonstrations.}
    \label{tab:suboptimal_grid}
    \begin{tabular}{c|c}
    \hline
        & Reward \\
        \hline
         GP-PTR & 7.42 $\pm$ 4.82\\
        \hline
         GP-ME-IRL & 3.31 $\pm$ 4.24\\
         \hline
         GP-LTR &  2.77 $\pm$ 4.30\\
         \hline
         T-REX & 0.42 $\pm$ 1.59\\
         \hline
         D-REX & 0.49 $\pm$ 2.10\\
         \hline
    \end{tabular}
\end{table}

In this sub-optimal exploratory demonstration case, policies obtained using the maximum entropy approach regularly fail, while the probabilistic temporal ranking continues to perform relatively well. Figure ~\ref{fig:suboptimal} shows a sample environment used for testing. The sub-optimal behaviour of the exploring model predictive control policies used for demonstration can result in frequent visits to undesirable states, which leads to incorrect reward attribution under a maximum entropy model. Probabilistic temporal ranking avoids this by using the looser assumption that states generally improve over time. T-REX performs extremely poorly here, as it is unable to learn from the limited number of demonstrations provided. D-REX also fails here, as it is unable to separate the uninformative exploratory portions of the demonstrating from the final exploitative portion of the demonstration policy. While D-REX works well for demonstrations that are sub-optimal due to noise, in this case of exploratory goal oriented tasks, the assumptions made by PTR are more applicable.

\begin{figure}
    \centering
    \includegraphics[width=0.45\textwidth]{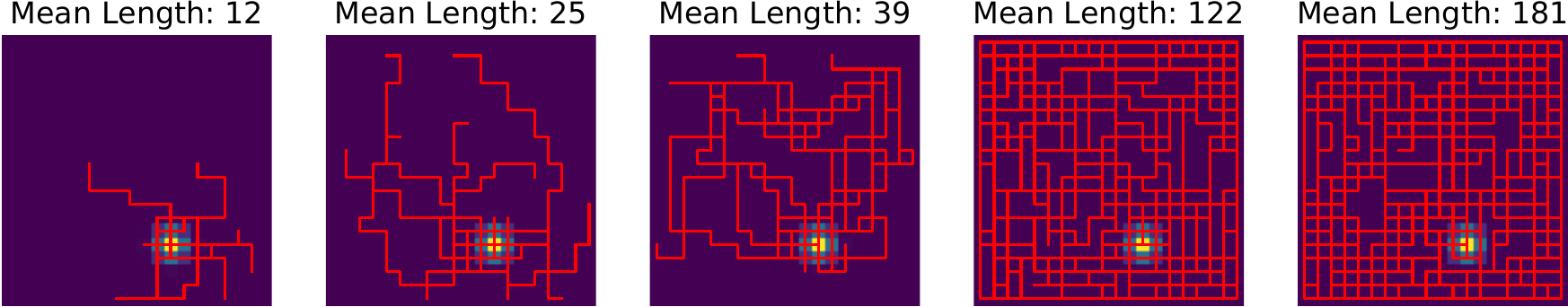}\\\vspace{2mm}
    \includegraphics[width=0.45\textwidth]{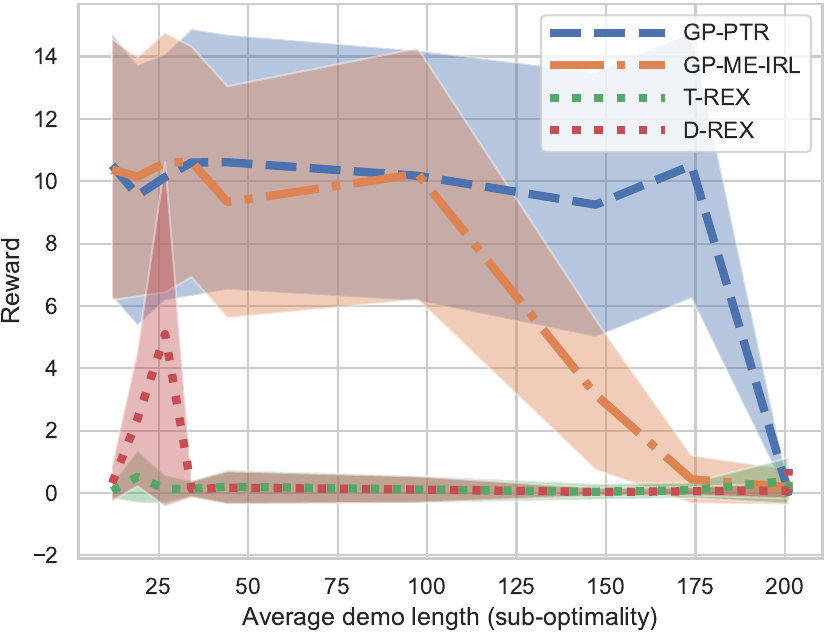}
    \caption{Policy returns using rewards learned with trajectories of increasing length show the degradation of GP-ME-IRL as trajectories become more exploratory. T-REX performs poorly here, as it needs both good and bad examples to learn a reward function. PTR performs well for both optimal and exploratory demonstrations.}
    \label{fig:optimality_len}
\end{figure}
Figure \ref{fig:optimality_len} shows the performance of PTR as trajectories become more exploratory. Here, 100 demonstrations were generated for a single environment and sorted by length. Reward models were then learned using subsets of 10 demonstrations of increasing length. Policies were trained to maximise these reward functions using value iteration. GP-ME-IRL rapidly degrades as trajectories become more exploratory. T-REX performs poorly here, as it needs both good and bad examples to learn a reward function. D-REX is unable to handle the exploratory trajectories. In contrast, PTR performs well for both optimal and exploratory demonstrations in these goal-oriented environments, failing only when demonstrations never reach the goal.
\begin{figure}
    \centering
    \includegraphics[width=0.45\textwidth]{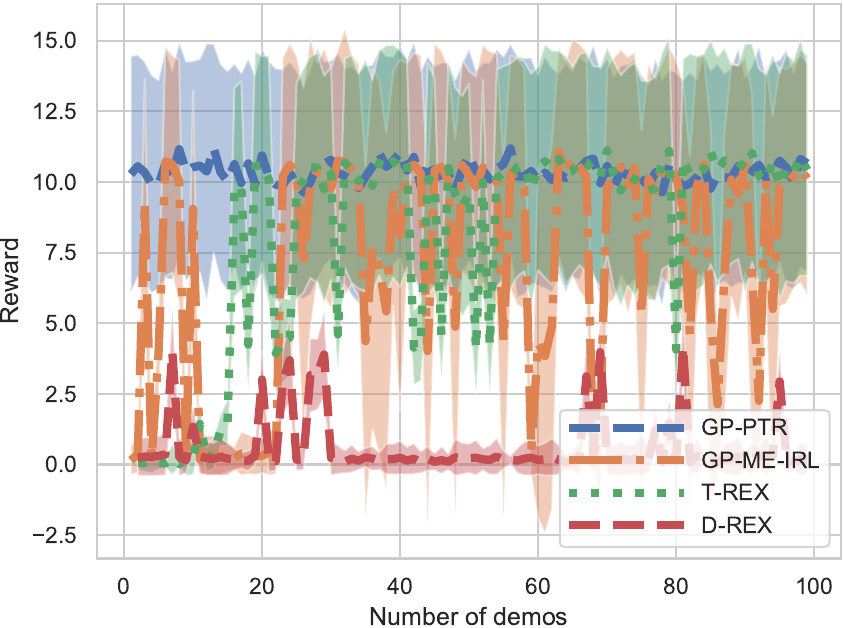}
    \caption{Policy returns using rewards learned with increasing numbers of demonstration trajectories of increasing length. T-REX starts to perform better with more demonstration data, while GP-ME-IRL is highly dependent on the quality of demonstrations. PTR performs well even with a limited number of demonstration sequences.}
    \label{fig:optimality_demos}
\end{figure}
Figure \ref{fig:optimality_demos} shows the rewards obtained by policies trained using rewards learned with increasing numbers of demonstrations. T-REX performs substantially better with more demonstration data and a good balance of optimal and exploratory trajectories, but struggles to learn from limited data. GP-ME-IRL and D-REX, which make similar assumptions about the reward, are highly dependent on the quality of demonstrations and thus extremely unreliable in this setting. In contrast, PTR performs well even with a limited number of demonstration sequences.

\subsection{A deterministic neural approximation to PTR}
\begin{figure}
    \centering
    \begin{overpic}[width=0.45\textwidth]{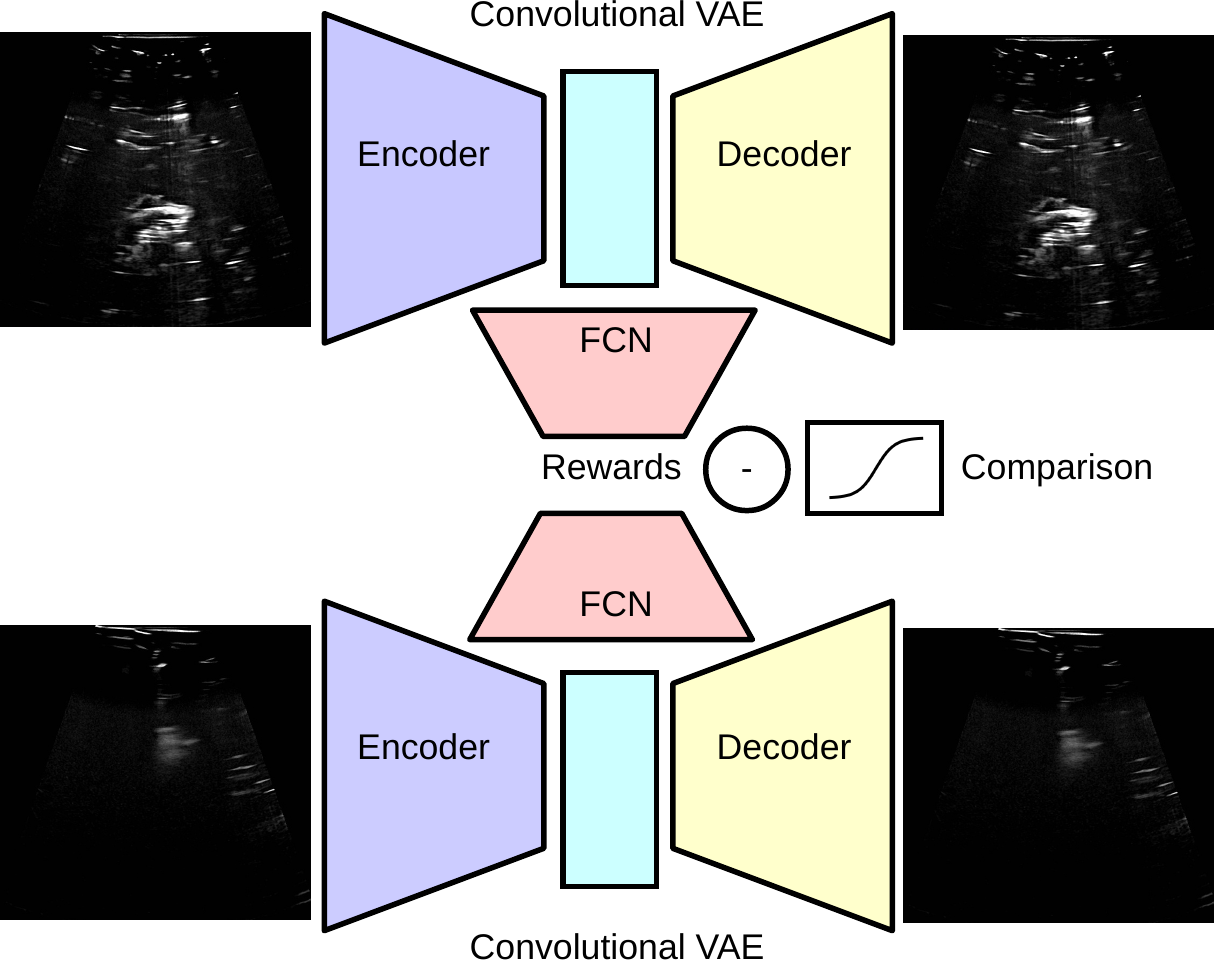}
    \put(47.75,65){\tiny{$p(\mathbf{x})$}}
    \put(59,62){\tiny{$p_\phi(\mathbf{Z}_t|\mathbf{x})$}}
    \put(30,13.5){\tiny{$q_\theta(\mathbf{x}|\mathbf{Z}_t)$}}
    \put(47.75,16){\tiny{$p(\mathbf{x})$}}
    \put(59,13.5){\tiny{$p_\phi(\mathbf{Z}_t|\mathbf{x})$}}
    \put(30,62){\tiny{$q_\theta(\mathbf{x}|\mathbf{Z}_t)$}}
    \put(70,34){\tiny{$h(g)$}}
    \put(47.5,33){\tiny{$r_\psi(\mathbf{x})$}}
    \put(47.5,47){\tiny{$r_\psi(\mathbf{x})$}}
    \put(84,36.5){\tiny{$g_i = 1$}}
    \put(5,78.5){\tiny{$\mathbf{Z}_{t_2}, t_2 > t_1$}}
    \put(5,29.5){\tiny{$\mathbf{Z}_{t_1}, t_1 \leq t_2$}}
    \end{overpic}
    \caption{Neural PTR approximation. Sampled images are auto-encoded, and a reward network predicts corresponding rewards, the sigmoid of the difference between these reward produces a comparison outcome probability. Weight sharing is indicated by colour. The network is trained jointly using a joint variational autoencoder and binary cross entropy loss. \label{fig:arch}}
\end{figure}

Given that learning from demonstration typically aims to require only a few trials, numerical inference under the fully Bayesian generative model described above is tractable, particularly if a sparse Gaussian process prior is used. However, in the case where a greater number of demonstrations or comparisons is available, we can approximate the fully probabilistic PTR model above with a deterministic model that can be trained in an end-to-end fashion, using the architecture in Figure~\ref{fig:arch}. Here, we replace the Gaussian process with a wide single layer fully connected network (FCN), $r_\psi(\mathbf{x})$, with parameters $\psi$, since single layer FCN's with i.i.d Gaussian weights are known to approximate a sample from a Gaussian processes \citep{neal1996priors} as model width tends to infinity. This approximation is trained by minimising a binary cross entropy loss over the expected comparison outcome alongside a variational autoencoder (VAE) objective,
\begin{eqnarray}
\mathcal{LL} &=& -\mathbb{E}_{\mathbf{x}_{t_1}\sim q_\theta}\left[\text{log} p_\phi(\mathbf{Z}_{t_1}|\mathbf{x})\right] + \mathbb{KL}\left(q_\theta(\mathbf{x}|\mathbf{Z}_{t_1})||p(\mathbf{x})\right) \nonumber\\
&-& \mathbb{E}_{\mathbf{x}_{t_2}\sim q_\theta}\left[\text{log} p_\phi(\mathbf{Z}_{t_2}|\mathbf{x})\right] + \mathbb{KL}\left(q_\theta(\mathbf{x}|\mathbf{Z}_{t_2})||p(\mathbf{x})\right) \nonumber\\&-& \frac{1}{N} \sum_{i=1}^N \left[ g_i\text{log}\left(h(g_i)\right) + (1-g_i)\text{log}\left(1-h(g_i)\right)\right]
\end{eqnarray}
using stochastic gradient descent. Here, $\mathcal{LL}$ denotes the overall loss, $p(\mathbf{x})$ is a standard normal prior over the latent space, $q_\theta(\mathbf{x}|\mathbf{Z}_{t_i})$ denotes the variational encoder, with parameters $\theta$, $p_\phi(\mathbf{Z}_{t_i}|\mathbf{x})$ represents the variational decoder, with parameters $\phi$, and $h(g)$ is the comparison output logit (sigmoid). $g_i$ is a temporal comparison outcome label, and $\mathbf{Z}_{t_i}$ denotes a training sample image, with $\mathbf{x}_t$ a sample from the latent space. Weight sharing is used for both the convolutional VAEs and FCNs. 

\begin{table*}[ht]
    \centering
    \caption{Returns after training for 500000 environment steps. 100 demonstrations are used for reward learning or imitation learning, and the best result across 3 seeds reported (100 test episodes). As expected, PTR performs well on goal-oriented tasks, but fails elsewhere. }\label{tab:gym}
\begin{tabular}{l|l||l||l}
\hline
& \textbf{Pendulum-v1} & \textbf{Hopper-v3} & \textbf{HalfCheetah-v3} \\ 
\hline
\textbf{Method} & Reward (Mean $\pm$  Std) & Reward (Mean $\pm$  Std)  & Reward (Mean $\pm$  Std) \\
\hline
Expert (PPO) & -150.4354 $\pm$ 88.1962 & 3437.0038 $\pm$ 8.1665 & 1193.6989 $\pm$ 62.7869\\
\hline
ML-PTR (PPO)& \textbf{-189.2376 $\pm$ 112.5798} & 183.4452 $\pm$ 1.8466 & 130.8885 $\pm$ 46.1780\\
\hline
GAIL & -315.1454 $\pm$ 193.8659 & \textbf{3415.7294 $\pm$ 1.9907} & \textbf{1220.1896 $\pm$ 92.7100}\\
\hline
AIRL & -948.4833 $\pm$ 107.4486 & 5.7232 $\pm$ 0.030 & 1000.2273 $\pm$ 63.5880\\
\hline
\hline
& \textbf{Acrobot-v1} & \textbf{Ant-v3} \\ 
\hline
\textbf{Method} & Reward (Mean $\pm$  Std) & Reward (Mean $\pm$  Std)\\
\hline
Expert (PPO) & -73.8100 $\pm$ 10.0297 & 800.8259 $\pm$ 85.1205 &\\
\hline
ML-PTR (PPO) & -81.5600 $\pm$ 21.4487 & -1558.8640 $\pm$ 2.1768\\
\hline
GAIL & -292.8000 $\pm$ 161.7387 & \textbf{886.3507 $\pm$ 25.8765} & \\
\hline
AIRL & \textbf{-77.8600 $\pm$ 19.3111} & -15.4500 $\pm$ 21.4068 &\\
\hline
\end{tabular}
\end{table*}

Once trained, the reward model is provided by encoding the input observation, and then predicting the reward using the FCN. This allows for rapid end-to-end training using larger datasets and gives us the ability to backpropagate the comparison supervisory signal through the autoencoder, potentially allowing for improved feature extraction in support of reward modelling. However, this comes at the expense of uncertainty quantification, which is potentially useful for the design of risk-averse policies that need to avoid regions of uncertainty. We investigate these trade-offs and the efficacy of the approximate model in more general reinforcement learning environments below, and in the context of autonomous ultrasound scanning in Section \ref{sec:Ultrasound}.

\subsection{General reinforcement learning environments}

It should be noted that while reward inference using probabilistic temporal ranking is capable of handling sub-optimal exploratory demonstrations in goal-oriented environments, this assumption does not hold for more general tasks. To illustrate this, we applied the deterministic neural PTR approximation above (ML-PTR) to a range of low dimensional continuous control environments\footnote{No autoencoder is used, as the regularisation provided is unnecessary for these low dimensional control problems.} \citep{gym}. Here, we collect 100 demonstrations from an agent trained using proximal policy optimisation (PPO) \citep{schulman2017proximal}, and infer rewards using ML-PTR. We then use the inferred reward function to train a PPO agent. We benchmark against AIRL \citep{fu2017learning} and GAIL \citep{ho2016generative}, popular imitation learning approaches for lower dimensional control tasks. We use the imitation toolbox \citep{gleave2022imitation} and Stable Baselines3 \citep{stable-baselines3} for these experiments.

As shown in Table \ref{tab:gym}, the policy trained using the ML-PTR reward performs best on goal oriented tasks (learning to balance Pendulum-v1 and swing up Acrobot-v1), but fails on the continuous control tasks (Ant-v3, Hopper-v3, HalfCheetah-v3), where the assumption of generally increasing reward over a demonstration does not hold.

\section{Autonomous ultrasound scanning}
\label{sec:Ultrasound}
For our primary experiment, we demonstrate the use of probabilistic temporal ranking in a challenging ultrasound scanning application. Here, we capture 10 kinesthetic demonstrations of a search for a target object using a compliant manipulator, and use only ultrasound image sequences (2D trapezoidal cross-sectional scans) to learn a reward model. Our goal is to use this reward model within a control policy that automatically searches for and captures the best image of a tumour-like mass\footnote{A roughly 30 mm x 20 mm blob of Blu tack original in a container of dimensions 200 mm x 150 mm x 150 mm.} suspended within a deformable imaging phantom constructed using a soft plastic casing filled with ultrasound gel.

This task is difficult because it involves a highly uncertain and dynamic domain. Obtaining stable ultrasound images requires contact with a deformable imaging phantom at an appropriate position and contact force, with image quality affected by the thickness of the ultrasound gel between the phantom and the probe, while air pockets within the phantom object can obscure object detection. Moreover, since the phantom deforms, air pockets and gel can move in response to manipulator contact. This means that kinesthetic demonstrations are inherently sub-optimal and exploratory, as they require that a demonstrator actively search for target objects, while attempting to locate a good viewpoint position and appropriate contact force. As in real-world medical imaging scenarios, the demonstrator is unable to see through the phantom object from above, so demonstrations are based entirely on visual feedback from an ultrasound monitor. 

\subsection{Active viewpoint selection}
\label{sec:GPMPC}
Although the proposed reward model can be used with any policy, we demonstrate its use by means of a Bayesian optimisation policy, selecting action $\hat{\mathbf{a}}_t$ that drives an agent to a desired end-effector position $\hat{\mathbf{s}}_t$, drawn from a set of possible states (a volume of acceptable end-effector positions) using an upper confidence bound objective function that seeks to trade off expected reward returns against information gain or uncertainty reduction.
 
Here, we learn a mapping between reward and end-effector positions using a surrogate Gaussian process model with a radial basis function kernel,
\begin{equation}
r_t \approx \mathcal{GP}\left(\mathbf{0},\text{RBF}(\mathbf{s}_t,l_p)\right)\label{eq:BO}.
\end{equation}
Length scale $l_p$ is determined by maximum likelihood estimation, using a search within the length scale bounds $l_p \in [1\mathrm{e}{-5}, 0.01]\,m$. Fixed measurement noise, $\alpha = 0.5$, is used when fitting the Gaussian process to account for the variability in reward that may be obtained in a given end-effector position, resulting from the variability introduced by contact with the deformable phantom, potential tumour motion and ultrasound gel spreading effects.

The Gaussian process is iteratively trained using a buffer of visited states (in our experiments these are 3D Cartesian end-effector positions) and the corresponding rewards predicted using the image-based reward model. Actions are then chosen to move to a desired state selected using the objective function,
\begin{equation}
   \hat{\mathbf{s}}_t = \text{argmax}_{\mathbf{s}_t} \hspace{2mm} \mu(\mathbf{s}_t) + \beta \sigma(\mathbf{s}_t). \label{eq:aq}
\end{equation}
Here $\mu(\mathbf{s}_t)$, $\sigma(\mathbf{s}_t)$ are the mean and standard deviation of the Gaussian process, and $\beta = 1$ is a hyperparameter controlling the exploration exploitation trade-off of the policy. This objective function is chosen in order to balance the competing objectives of visiting states that are known to maximise reward with gaining information about values of states for which the model is more uncertain. In our ultrasound imaging application, actions are linear motions to a desired Cartesian state.
\begin{figure*}
\centering
\includegraphics[width=0.95\textwidth]{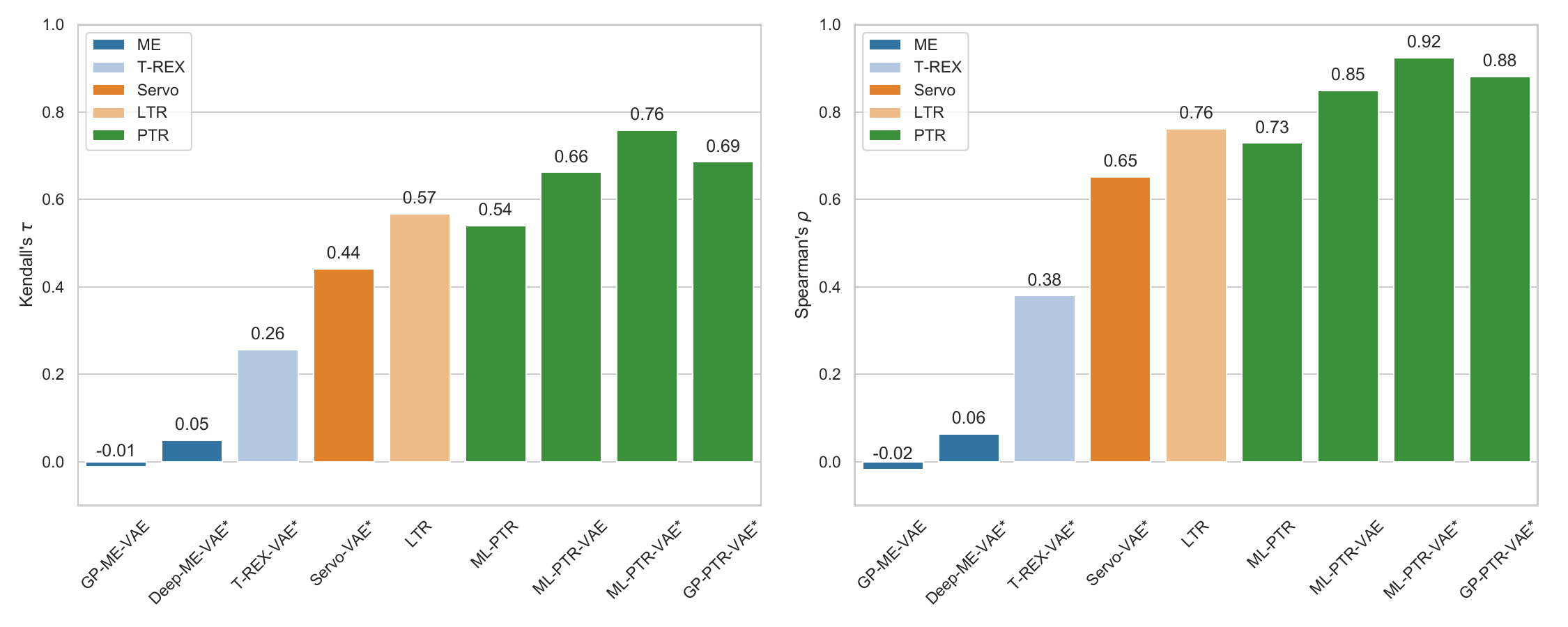}
\caption{Reward model association with human image ratings shows that temporal ranking (green) reward inference models strongly agree with human preferences. A Spearman rank correlation of $\rho=1$ indicates an identical rank or ordering, while $\rho=-1$ indicates a completing opposing order. Similarly, Kendall's $\tau=1$ indicates that the relative rank assigned to images is identical to that assigned by the human annotator, while $\tau=-1$ would indicate opposing ranks. \label{fig:quantitative_comparison}}
\end{figure*}

Since the GP starts with no prior information about reward, and is re-fit online after each position is visited, the Bayesian optimisation policy naturally transitions from exploration and becomes more exploitative as additional information is gained. Note that, for the ultrasound case, no policy is ever `trained', instead we optimise the learned reward function online for each new environment using the fixed Bayesian optimisation strategy.

It should also be noted that any policy can be used to optimise rewards predicted using probabilistic temporal ranking. We selected a Bayesian optimisation strategy for online experiments due to its prevalence in active viewpoint selection literature, and because of its ability to deal with uncertainty in state rewards arising the dynamic structure of the deformable imaging phantom.

\subsection{Reward inference evaluation}

Figure \ref{fig:unseen_reward} shows predicted reward sequences for sample expert demonstration traces held out from model training. It is clear that the ranking reward model captures the general improvements in image quality that occur as the demonstrator searches for a good scanning view, and that some searching is required before a good viewpoint is found. Importantly, the slack in the pairwise ranking model, combined with the model assumption that similar images result in similar rewards, allows for these peaks and dips in reward to be modelled, as probabilistic temporal ranking does not assume monotonically increasing rewards.  

We qualitatively assessed the image regions and features identified using the reward model using saliency maps (Figure \ref{fig:saliency}), which indicated that the proposed approach has learned to associate the target object with reward. 

In order to quantitatively evaluate the performance of probabilistic temporal ranking for autonomous ultrasound imaging, approximately 5000 ultrasound images from a set of 10 demonstration sequences were ordered in terms of human preference by collecting 5000 human image comparison annotations and applying the ranking model of \citep{burke2017rapid}. We evaluate reward inference models in terms of how well they agree with this human labelling using Kendall's $\tau$, a measure of the ordinal association between observation sets, and Spearman's $\rho$, a measure of rank correlation. Figure \ref{fig:quantitative_comparison} shows these results.

We benchmark probabilistic temporal ranking (PTR) against a maximum entropy reward model with both a Gaussian process prior (GP-ME-VAE*) \citep{levine2011nonlinear} and a neural network prior (Deep-ME-VAE*) \citep{wulfmeier2015maximum}, a monotonically increasing linear temporal ranking model (LTR) \citep{angelov2019composing}, T-REX-VAE* \citep{brown2019extrapolating} and a servoing reward model (Servo-VAE*) based on the cosine similarity of a latent image embedding to a final image captured. As in the grid world experiments, we provide T-REX with labelling information using trajectory length as a heuristic for demonstration quality. We also include a number of ablation results for probabilistic temporal ranking models. Model parameters are provided in the appendices.

Here, VAE* denotes the use of pre-trained image embedding learned independently using variational autoecoding, ML-PTR refers to a model trained without decoding the latent embedding (no autoencoder loss), ML-PTR-VAE refers to a maximum likelihood model trained jointly with both a variational autoencoding and pairwise ranking objective, and GP-PTR-VAE* denotes the use of the probabilistic temporal ranking with a pre-trained image embedding. It should be noted that all reward models were inferred without policy search, by directly optimising the reward objective.

\begin{figure*}
    \centering
    \subfloat[Probabilistic temporal ranking reward (Average human image rating: $0.254 \pm 0.079$)]{
    \includegraphics[width=0.85\textwidth]{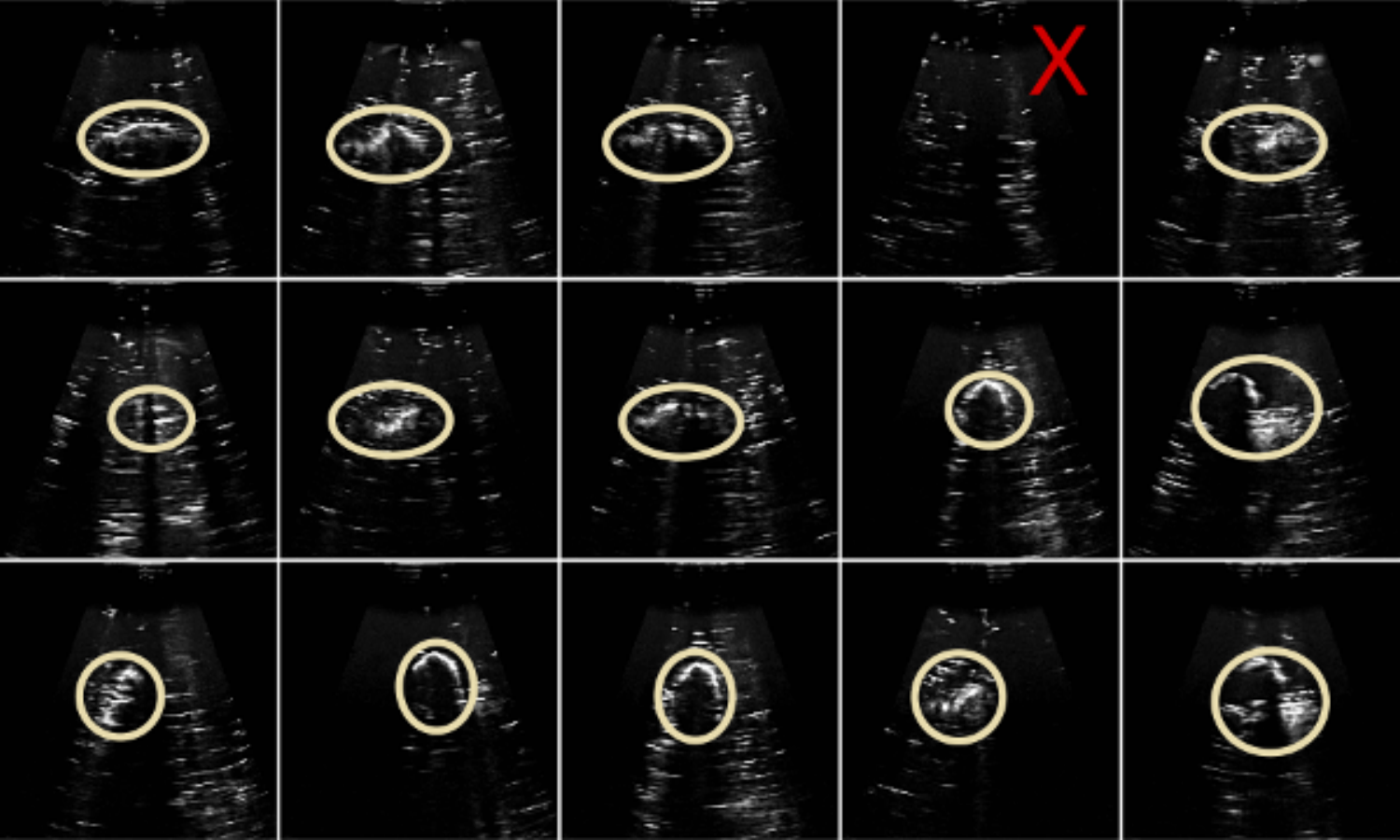}
    }\\
   \subfloat[Maximum entropy reward (Average human image rating: $0.119 \pm 0.198$)]{
    \includegraphics[width=0.85\textwidth]{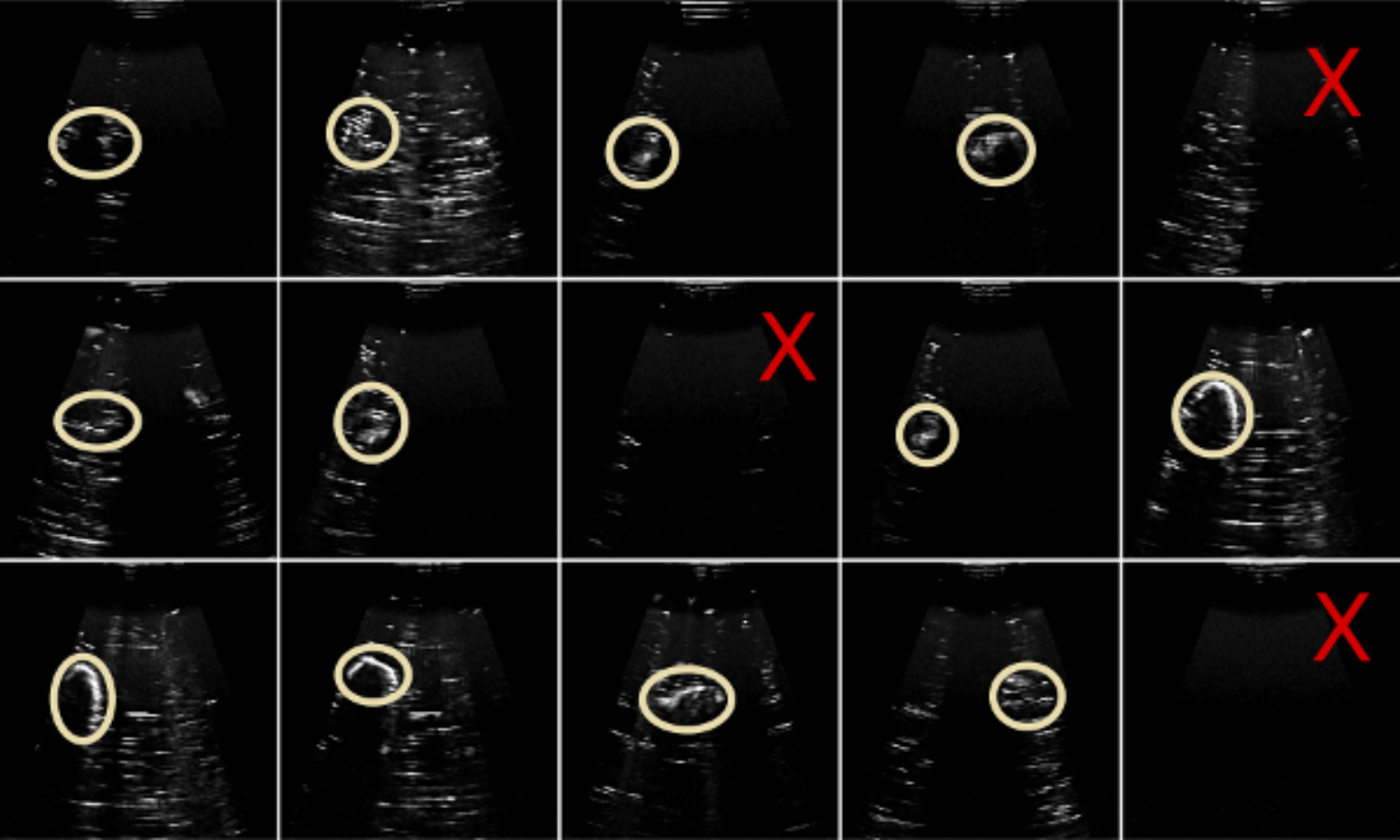}
    }
    \caption{Final images obtained after policy convergence clearly show that images obtained using probabilistic temporal ranking are much clearer and capture the target object far more frequently than the maximum entropy reward. Target objects are circled, failures marked with a cross. (Images are best viewed electronically, with zooming. See companion site, \href{https://sites.google.com/view/ultrasound-scanner}{\url{https://sites.google.com/view/ultrasound-scanner}}, for higher resolution images.) \label{fig:final_ims}}
\end{figure*}

It is clear that PTR outperforms baseline approaches. The maximum entropy reward models fail to learn adequate reward models. Servoing proved somewhat effective, but is unlikely to scale to more general problems and use cases. PTR improves upon LTR, illustrating the importance of allowing for non-monotonically increasing rewards. T-REX performs much better than maximum entropy approaches, but does not recover the underlying reward function. This is most likely due to the limited number of demonstrations available, but also potentially due to the fact that trajectory length is not always indicative of scan quality in this setting. In some demos, a set of very good quality images could be obtained after extensive exploration, while in others, a passable set of scans may have been obtained after relatively little searching. As a result, there is no clear or objective measure of the quality of a scanning sequence suitable for use with supervised learning approaches like T-REX.

The ablation results show that variational autoencoding produces better reward models. This is most likely due to its regularising effect, which helps to avoid over-fitting to insignificant image appearance differences. Directly optimising without this regularising effect (ML-PTR) essentially results in a monotonically increasing reward model, and produces similar results to LTR. Interestingly, learning an independent auto-encoding and using a single layer bottleneck reward network or Gaussian process, proved to be an extremely effective strategy.

\begin{figure*}[!ht]
    \centering
    \includegraphics[width=\textwidth]{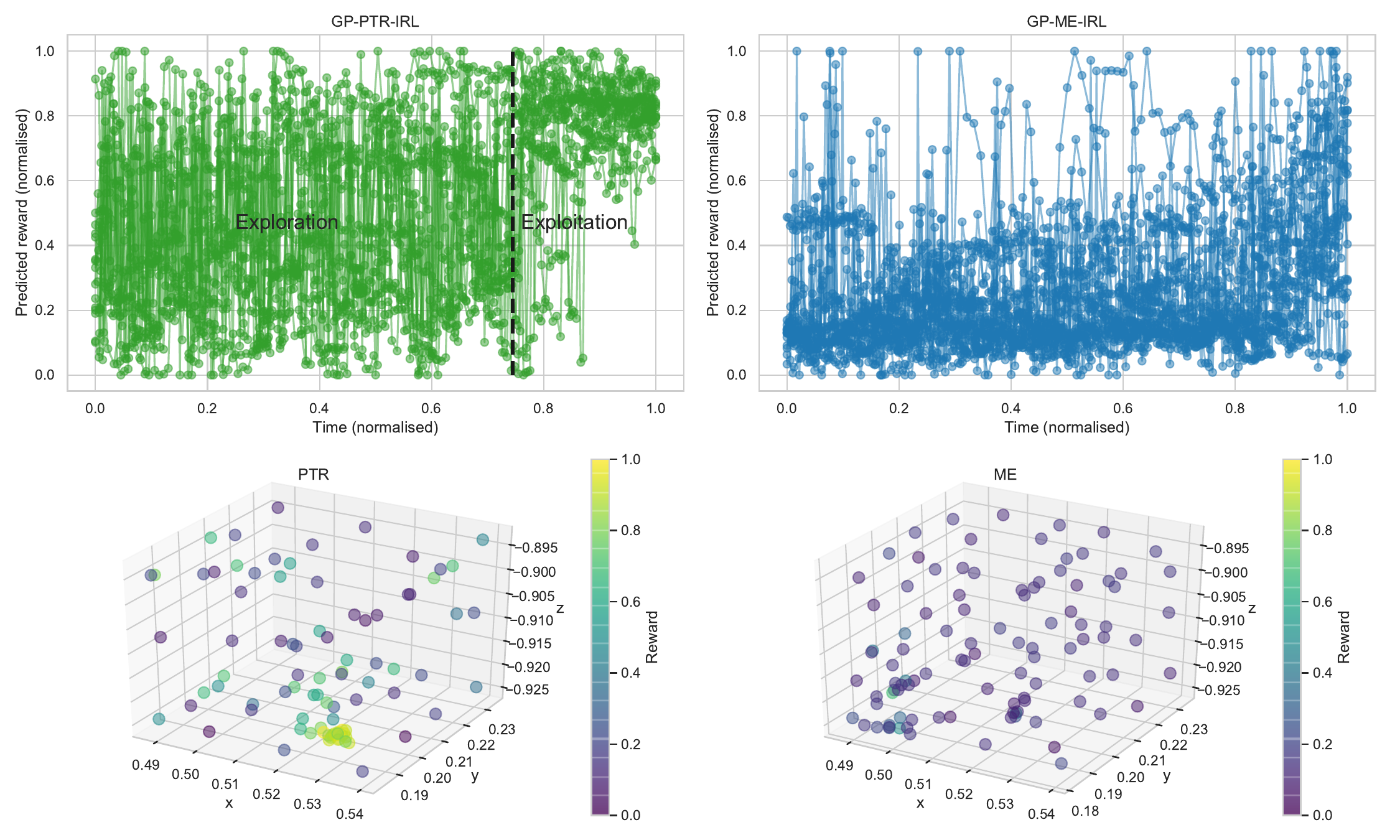}
    \caption{Reward traces (top) show that the probabilistic temporal ranking reward is stable enough for the BO robot policy to explore the volume of interest (varying reward) before exploiting (stable reward). The maximum entropy reward is extremely noisy, indicating that it has failed to consistently associate high quality ultrasound image features with reward. This can also be observed when the 3D positions selected by the Bayesian optimisation policy are visualised (bottom), and coloured by the reward associated with the ultrasound images obtained when visiting these locations. The PTR policy first explores the allowable search region, before converging to an optimal viewing position (the cluster of high reward points). A policy using the maximum entropy reward model fails to locate the target object. \label{fig:reward_traces}}
\end{figure*}

We believe that maximum entropy reward inference fails for two primary reasons. First, probabilistic temporal ranking produces substantially more training data, as each pair of images sampled (50 000 pairs) from a demonstration provides a supervisory signal. In contrast, the maximum entropy approach treats an entire trajectory as a single data point (10 trajectories), and thus needs to learn from far fewer samples, which is made even more challenging by the high dimensional image inputs. Secondly, the maximum entropy reward assumes that frequently occurring features are a sign of a good policy, which means that it can mistakenly associate undesirable frames seen during the scan's searching process for frames of high reward.

\section{Policy evaluation}

For policy evaluation, we compare probabilistic temporal ranking with a Gaussian process maximum entropy inverse reinforcement learning approach. For both models we use the same latent feature vector (extracted using a stand-alone variational autoencoder following the architecture in Figure \ref{fig:arch}), and the same Bayesian optimisation policy to ensure a fair comparison. 

We compare the two approaches by evaluating the final image captured during scanning, and investigating the reward traces associated with each model.

Trials were repeated 15 times for each approach, alternating between each, and ultrasound gel was replaced after 10 trials. Each trial ran for approximately 5 minutes, and was stopped when the robot pose had converged to a stable point, or after 350 frames had been observed. A high quality ultrasound scan is one in which the contours of the target object stand out as high intensity, where the object is centrally located in a scan, and imaged clearly enough to give some idea of the target object size (see Figure \ref{fig:expert}). 

As shown in Figure \ref{fig:final_ims}, the probabilistic temporal ranking model consistently finds the target object in the phantom, and also finds better rated images. Mean and standard deviations in image ratings were obtained using the rating model (see above) trained for reward evaluation using human image preference comparisons. The maximum entropy approach fails more frequently than the ranking approach, and when detection is successful, tends to find off-centre viewpoints, and only images small portions of the target object. 

It is particularly interesting to compare the reward traces for the probabilistic temporal ranking model to those obtained using maximum entropy IRL when the Bayesian optimisation scanning policy is applied. Figure \ref{fig:reward_traces} overlays the reward traces obtained for each trial. The maximum entropy reward is extremely noisy throughout trials, indicating that it has failed to adequately associate image features with reward. Similar images fail to consistently return similar rewards, so the Bayesian optimisation policy struggles to converge to an imaging position with a stable reward score. In contrast, the reward trace associated with the probabilistic temporal ranking contains an exploration phase where the reward varies substantially as the robot explores potential viewpoints, followed by a clear exploitation phase where an optimal viewpoint is selected and a stable reward is returned.

\begin{figure}
    \centering
    \begin{tabular}{cc}
    \includegraphics[height=0.156\textwidth]{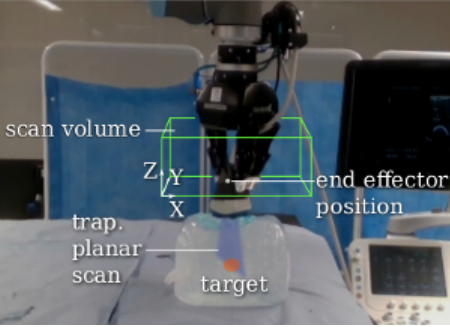} &
    \includegraphics[height=0.156\textwidth]{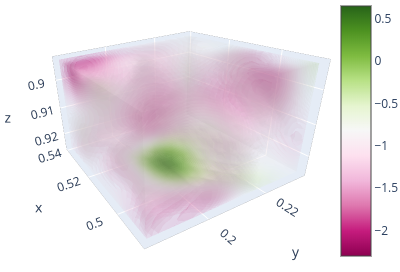} \\
    (a) Scan volume  & (b) Reward map
    \end{tabular}
    \caption{A visualisation of the reward map (b) inferred by the Bayesian optimisation Gaussian process during scanning shows that it attributes high rewards (green) when the probe is pressed against the container directly above the target. The scan volume, or the support of the reward map, is illustrated using a green wireframe in the setup (a). \label{fig:volume}}
\end{figure}

Figure \ref{fig:volume} shows the predicted reward over the search volume (a 50 mm x 50 mm x 30 mm region above the imagining phantom) for a PTR trial, determined as part of the Bayesian optimisation search for images with high reward, from reward and position samples (see Figure \ref{fig:reward_traces}). Here, we capture images at 3D end-effector locations according to the Bayesian optimisation policy, and predict the reward over the space of possible end-effector states using (\ref{eq:BO}). Importantly, the Gaussian process proxy function is able to identify an ultrasound positioning region associated with high reward. This corresponds to a position above the target object, where the contact force with the phantom is firm enough to press through air pockets, but light enough to maintain a thin, air-tight layer of gel between the probe and phantom.

\section{Conclusion} 
\label{sec:conclusion}

This work introduced probabilistic temporal ranking, an approach to reward inference from exploratory demonstrations for visual servoing or active viewpoint selection tasks. Here, we take advantage of the fact that exploratory demonstrations, whether optimal or sub-optimal, often involve steps taken to improve upon an existing state. Results show that leveraging this to infer reward through a ranking model is more effective than common IRL methods in exploratory cases where demonstrations require a period of discovery in addition to reward exploitation and when observation traces are high dimensional.

This paper also shows how the proposed reward inference model can be used for a challenging ultrasound imaging application. Here, we learn to identify image features associated with target objects using kinesthetic scanning demonstrations that are exploratory, as they inevitably require a search for an object and position or contact force that returns a good image. Using this within a policy that automatically searches for positions\footnote{For videos and higher resolution scan images, along with post publication links to code see \url{https://sites.google.com/view/ultrasound-scanner} } and contact forces that maximise a learned reward, allows us to automate ultrasound scanning.

When comparing with human scanning, a primary challenge we have yet to overcome is that of spreading ultrasound gel smoothly over a surface. Human demonstrators implicitly spread ultrasound gel evenly over a target as part of the scanning process so as to obtain a high quality image. The Gaussian process policy used in this work is unable to accomplish this, which means scans are still noisier than those taken by human demonstrators. Moreover, human operators typically make use of scanning parameters like image contrast, beam width and scanning depth, which we kept fixed for these experiments. Nevertheless, the results presented here show extensive promise for the development of targeted automatic ultrasound imaging systems, and open up new avenues towards semi-supervised medical diagnosis.

\appendix
\section*{Appendices}

\subsection*{Hyper-parameter priors}

\begin{figure}[!ht]
    \centering
    \includegraphics[width=0.23\textwidth]{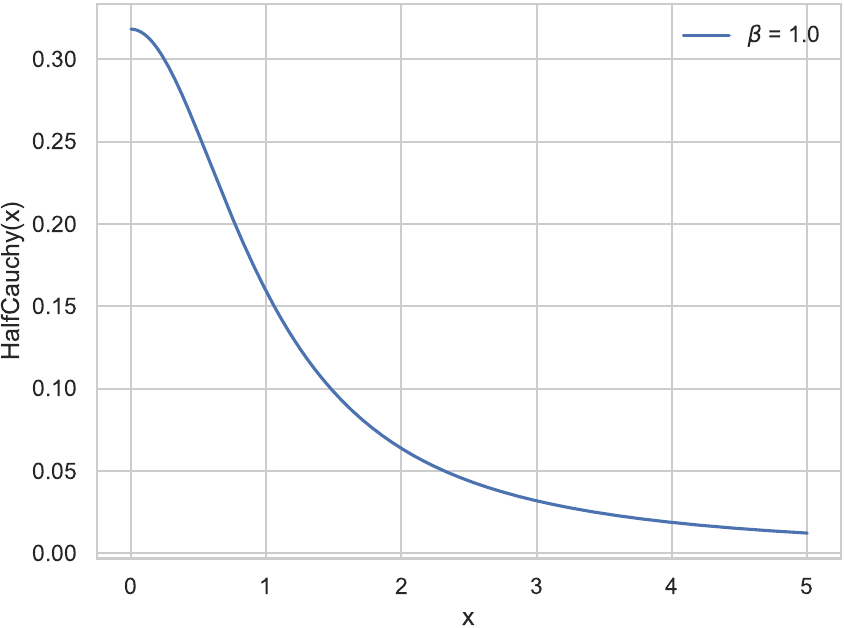}\includegraphics[width=0.23\textwidth]{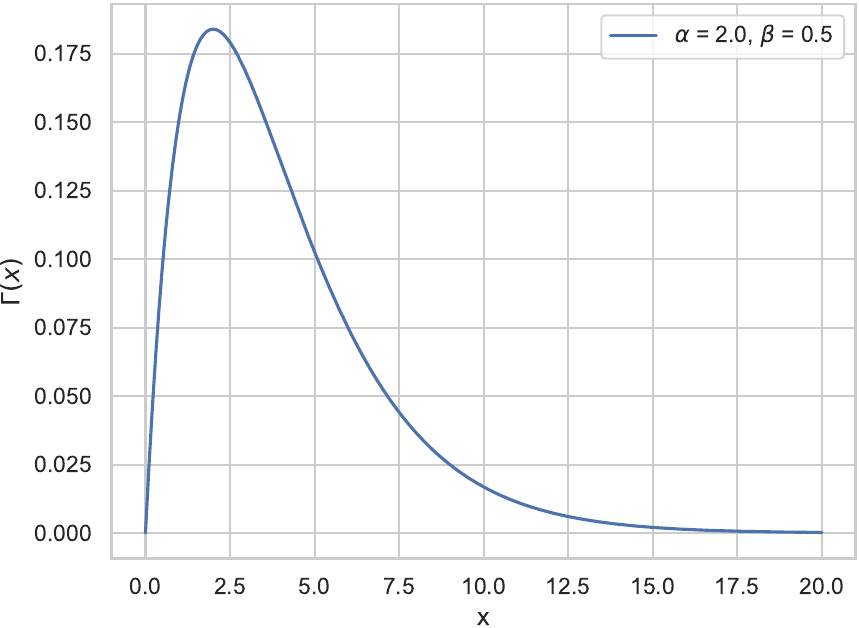}
    \caption{Prior distributions used for PTR hyper-parameters.\label{fig:Hypers}}
\end{figure}

\subsection*{Network Architectures}
Table \ref{tab:CVAE} shows the parameters and training settings used for the convolutional neural architecture experiments.
\begin{table}[!ht]
    \centering
    \caption{Neural architecture parameters\label{tab:CVAE}}
    \begin{tabular}{|l|c|}
    \hline
    Convolutional VAE &\\
    \hline
    Batch size & 128\\
    Training epochs & 100\\
    Adam optimiser & learning rate$=1e-4$\\
    Input dims &  $112 \times 112 \times 1 \in (0,1) $\\
    \hline
    Encoder &\\
    \hline
     Conv 32 & $5 \times 5$ kernel, relu, strides 2 \\     
     Conv 64 & $5 \times 5$ kernel, relu, strides 2 \\
     Conv 128 & $5 \times 5$ kernel, relu, strides 2 \\
     Conv 256 & $5 \times 5$ kernel, relu, strides 2 \\
     Dense FC & $1024$ neurons, relu \\
     Dense FC & $16 \times 2$ output (mean, variance)\\ 
     \hline
     Decoder &\\
    \hline
     Dense FC & $1024$ neurons, relu \\
     Conv transpose 128 & $5 \times 5$ kernel, relu, strides 2 \\     
     Conv transpose 64 & $5 \times 5$ kernel, relu, strides 2 \\
     Conv transpose 32 & $6 \times 6$ kernel, relu, strides 2 \\
     Conv transpose 1 & $6 \times 6$ kernel, relu, strides 2 \\
     Output dims &  $112 \times 112 \times 1 \in (0,1) $\\
     \hline
     Reward predictor & \\
     \hline
     Dense FC & relu, output dims 1 \\
     \hline
    \end{tabular}
\end{table}

Table \ref{tab:FC} shows the parameters and training settings used for the low-dimensional neural architecture experiments.
\begin{table}[!ht]
    \centering
    \caption{Neural architecture parameters (OpenAI Gym Environments)\label{tab:FC}}
    \begin{tabular}{|l|c|}
    \hline
    RewardNet &\\
    \hline
    Batch size & 64\\
    Training epochs & 1000\\
    AdamW optimiser & learning rate$=1e-4$\\
    \hline
    Encoder &\\
    \hline
    RunningNorm &\\
     Dense FC & $32 \times 32 $, relu,\\     
     Dense FC & $32 \times 32 $, relu,\\    
     \hline
     Reward predictor &\\
     \hline
     Dense FC & $32 \times 1 $, relu \\ 
     \hline
    \end{tabular}
\end{table}

\subsection*{Learning to play Atari Breakout}

Although this paper has primarily explored PTR in the context of exploratory demonstrations, PTR is also of use in a wide range of visual IRL tasks, particularly open ended `survival' settings. In these more general cases, PTR rewards image features corresponding to time spent in an environment, and becomes more akin to an intrinsic motivation strategy \citep{barto2013intrinsic}.  

\begin{figure}
    \centering
    \includegraphics[width=0.45\textwidth]{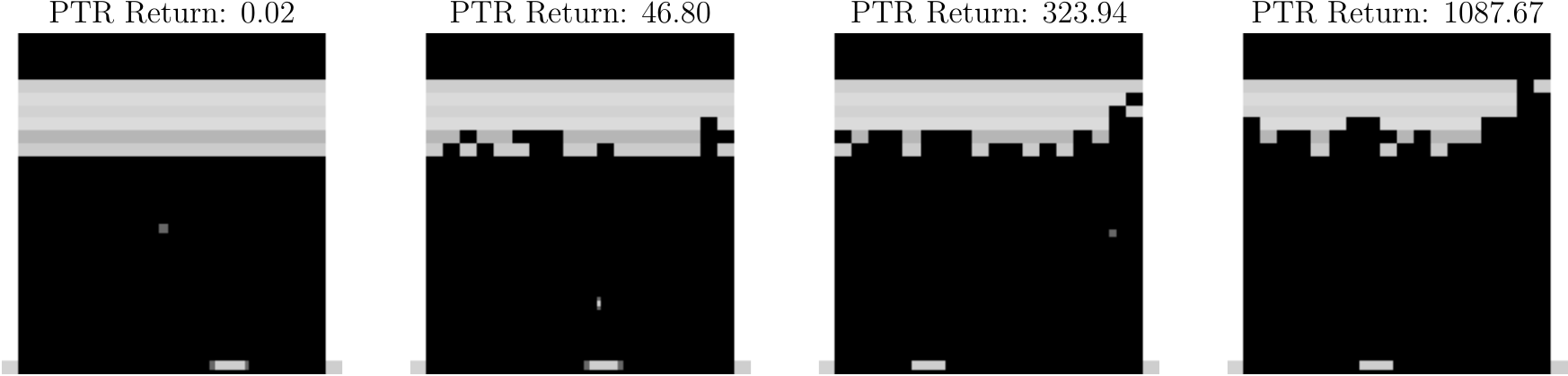}\\
    \vspace{5mm}
    \includegraphics[width=0.45\textwidth]{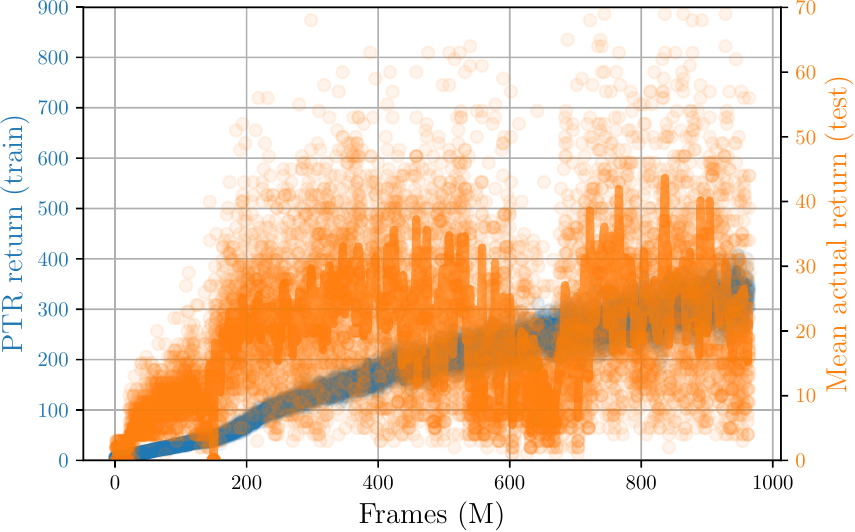}
    \caption{Top: Game states associated with PTR rewards. Bottom: Returns (cumulative reward) obtained for an A3C policy trained using PTR. The blue axes and curve provides the increase in return as a function of frames seen during training, while the orange curve provides the corresponding true reward obtained by the trained policy. Axes are aligned by linearly scaling by the ratio of the maximum reward inferred using PTR and the maximum true reward of an A3C policy used to gather demonstrations. \label{fig:breakout}}
\end{figure}

This is illustrated below using the Atari game Breakout, where a PTR reward function is learned from 20 demonstrations obtained by an agent trained using A3C\footnote{\href{https://github.com/greydanus/baby-a3c}{\url{https://github.com/greydanus/baby-a3c}}} \citep{mnih2016asynchronous}. The PTR reward function was then used to train a second agent using A3C, which, as illustrated in Figure \ref{fig:breakout}, learns to play Breakout reasonably well. 

\begin{acknowledgements}
We are particularly grateful to the Edinburgh RAD group and Dr Paul Brennan for valuable discussions and recommendations.
\end{acknowledgements}

%
\section*{Conflict of interest}

S. Ramamoorthy is vice president at five.ai, an autonomous driving company whose focus lies outside the domain of this paper. The remaining authors confirm that no other conflict of interest exists.

This work was supported by funding from the Turing Institute, as part of the Safe AI for surgical assistance project. K. Subr was supported by a Royal Society URF and K. Lu was partly supported by an  EPSRC grant: Honey Sampling.

\bibliographystyle{spbasic}      
\bibliography{references}   

\end{document}